%% file: main.tex
\documentclass[10pt,twocolumn,letterpaper]{article}

\usepackage[pagenumbers]{cvpr}              
\usepackage{graphicx}
\usepackage[graphicx]{realboxes}
\usepackage{amssymb}
\usepackage{colortbl}
\usepackage{multirow}
\usepackage{xcolor}
\usepackage{subcaption}
\usepackage{bm}
\usepackage{float, placeins}

\input{preamble}

\definecolor{cvprblue}{rgb}{0.21,0.49,0.74}
\usepackage[pagebackref,breaklinks,colorlinks,allcolors=cvprblue]{hyperref}

\def\paperID{} 
\def\confName{CVPR}
\def\confYear{2026}

\title{SimLBR: Learning to Detect Fake Images by Learning to Detect Real Images}

\author{Aayush Dhakal$^1$ \quad
Subash Khanal$^1$ \quad
Srikumar Sastry$^1$ \quad
Jacob Arndt$^2$ \quad
Philipe Ambrozio Dias$^2$ \\ 
Dalton Lunga$^2$ \quad
Nathan Jacobs$^1$\\[\bigskipamount]
Washington University in St. Louis$^1$ \quad Oak Ridge National Laboratory$^2$\\
 }
 
\begin{document}

\maketitle
\input{sec/0_abstract}    
\input{sec/1_intro}

\input{sec/related}
\input{sec/2_method}
\input{sec/3_results}

\input{sec/conclusion}

{
    \small
    \bibliographystyle{ieeenat_fullname}
    \bibliography{main}
}
\input{sec/X_suppl}


\end{document}

%% file: preamble.tex
\usepackage{pifont}
\newcommand{\xmark}{\ding{55}}
\newcommand{\chk}{\ding{51}}
\definecolor{darkgreen}{RGB}{0,100,0}
\definecolor{tblue}{HTML}{2E86AB}



%% file: sec/0_abstract.tex
\begin{abstract}
The rapid advancement of generative models has made the detection of AI-generated images a critical challenge for both research and society. Recent works have shown that most state-of-the-art fake image detection methods overfit to their training data and catastrophically fail when evaluated on curated hard test sets with strong distribution shifts. In this work, we argue that it is more principled to learn a tight decision boundary around the real image distribution and treat the fake category as a sink class. To this end, we propose SimLBR, a simple and efficient framework for fake image detection using Latent Blending Regularization (LBR). Our method significantly improves cross-generator generalization, achieving up to +24.85\% accuracy and +69.62\% recall on the challenging Chameleon benchmark. SimLBR is also highly efficient, training orders of magnitude faster than existing approaches. Furthermore, we emphasize the need for reliability-oriented evaluation in fake image detection, introducing risk-adjusted metrics and worst-case estimates to better assess model robustness. All code and models will be released on HuggingFace and GitHub.    
    
\end{abstract}

%% file: sec/1_intro.tex
\section{Introduction}
\label{sec:intro}

\begin{figure}[!t] 
  \centering
  \begin{subfigure}[t]{0.48\linewidth}
    \centering
    \includegraphics[width=\linewidth]{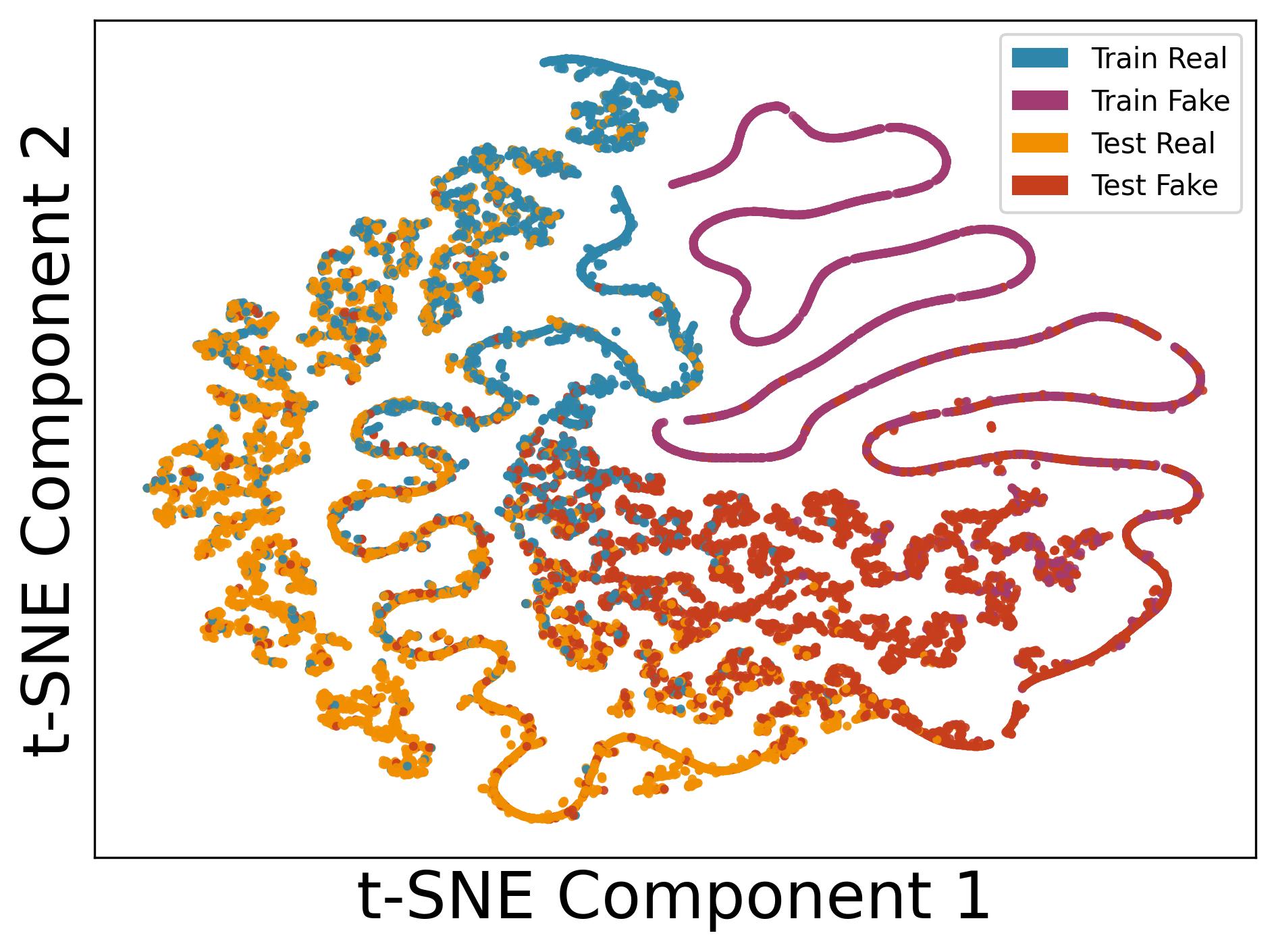}
     \subcaption{w/o LBR}
    \label{fig:wukong_tsne_a}
  \end{subfigure}\hfill
  \begin{subfigure}[t]{0.48\linewidth}
    \centering
    \includegraphics[width=\linewidth]{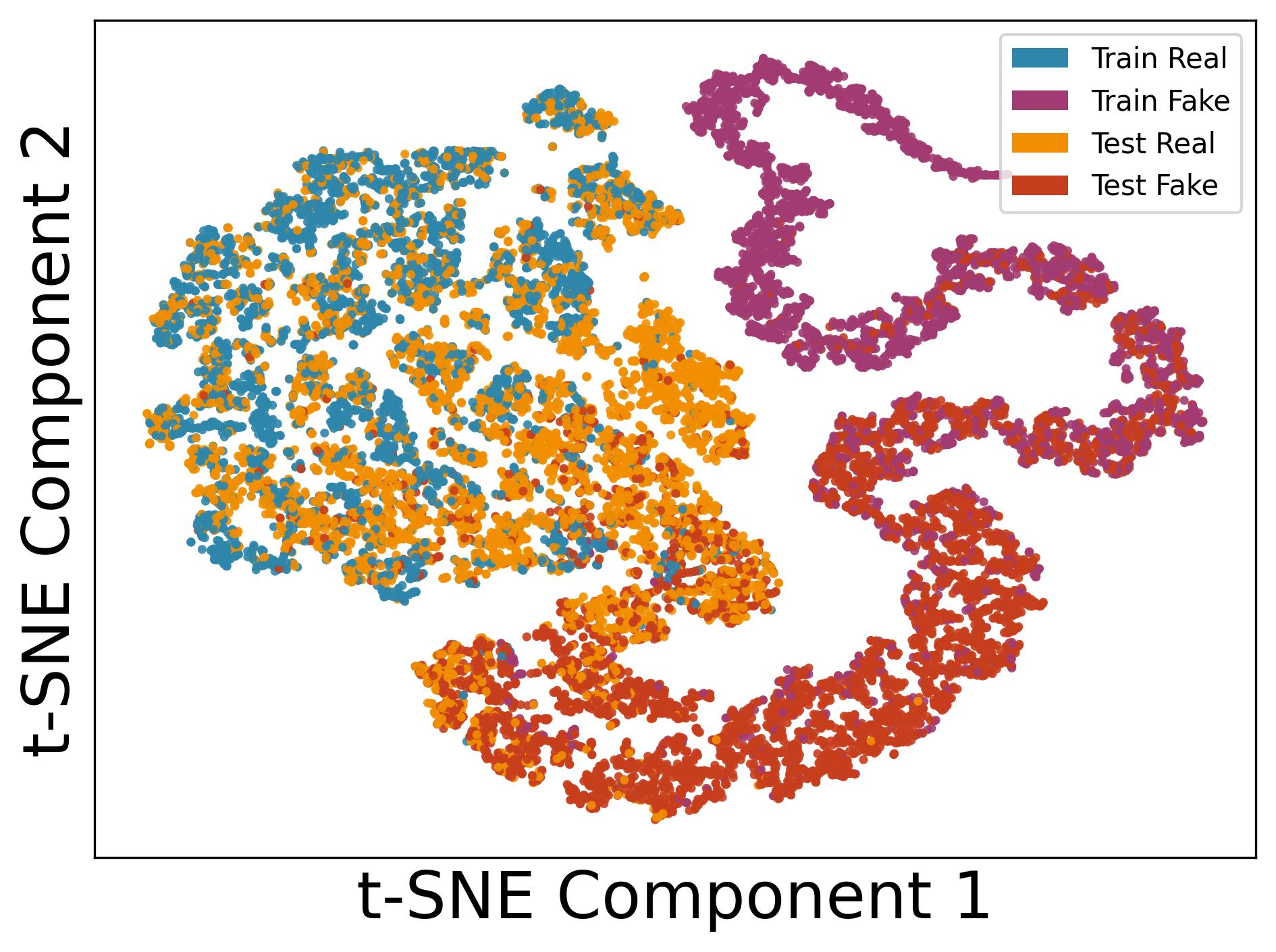}
     \subcaption{w/ LBR}
    \label{fig:wukong_tsne_b}
  \end{subfigure}
  \caption {\textbf{t-SNE Visualization of Latent Blending Regularization: }Prior methods tend to overfit to generator-specific artifacts, leading to fake images from unseen models being misclassified as real. Incorporating LBR encourages the detector to learn a tighter boundary around the real image distribution (\textcolor{tblue}{blue}/\textcolor{orange}{orange} cluster), ensuring that any sample lying outside this region—regardless of its generative source—is correctly identified as fake.}
  \label{fig:wukong_tsne}
\end{figure}

The quality of AI-generated imagery has advanced dramatically in recent years, driven by the rapid evolution of generative models. Today’s state-of-the-art systems~\cite{midjourney, esser2024scaling, nano} can synthesize photorealistic images that are often indistinguishable from real ones, posing serious challenges to information integrity, media authenticity, and public trust. As a result, developing robust and reliable AI-generated image detection models has become an urgent research priority.

While it is relatively easy to train a detector that performs well on fakes produced by a specific generative model, real-world applications demand detectors that can generalize to unseen or more advanced models~\cite{ojha2023towards,yansanity,zhong2023patchcraft}. Unfortunately, recent studies have shown that many existing detectors tend to overfit, learning superficial cues specific to the training data rather than generalizable features. Consequently, their performance collapses when evaluated on images synthesized by novel generators, rendering them unreliable in real safety-critical scenarios. 

This behavior was analyzed in detail by ~\citet{ojha2023towards}, who showed that during training, most detectors learn to identify generator-specific fingerprints rather than modeling the underlying distributional differences between real and fake images. As a result, these models fail to generalize and instead rely on superficial artifacts present in the training generator. When such artifacts are absent in new generative models, detectors tend to misclassify fake images as real, effectively treating the real category as a \textit{sink class}~\cite{ojha2023towards}. In this context, a sink class refers to a class that absorbs all out-of-distribution samples, regardless of their true label.

To prevent overfitting to generator-specific artifacts, several prior works~\cite{nguyen2024laa, Shiohara_2022_CVPR, tian2024real} avoid using fake images altogether and train detectors using only real images. These approaches construct pseudo-fake samples by applying hand-crafted manipulations to real images and then learn a decision boundary between real and pseudo-fake examples. Although conceptually appealing, this strategy has notable limitations. First, these methods operate in the pixel space, which often requires task-specific preprocessing (for example, facial landmark detection) to produce meaningful pseudo-fakes, making the approach restrictive and largely confined to domains such as facial forgery detection. Second, the pseudo-fake samples are generated using a narrow set of low-level image perturbations, which introduce only a narrow set of artifacts during training. As a result, the detector may still learn decision boundaries tied to a limited family of artifacts rather than a generalizable separation between real and fake images, leaving it vulnerable to failure when encountering fakes from more advanced generators.

UnivFD~\cite{ojha2023towards} proposed that learning a detector using a pretrained latent space is better, as it prevents the model from optimizing to detect artifacts directly in the pixel space. However, this approach is still insufficient to truly generalize to unseen generative models. This behavior was particularly highlighted in the recent work AIDE~\cite{yansanity}, where they argued that current datasets for evaluating AI-generated image detection models (e.g. AIGCDetect~\cite{wang2020cnn} and GenImage~\cite{zhu2023genimage}) are insufficient as they use randomly generated images with no guarantees of perceptual quality. Evaluating on these benchmarks might result in an inflated confidence on the true performance of a given model. To this end, they proposed a new dataset, Chameleon, which contains fake images from a diverse set, where each sample passes a human perception \textit{Turing Test}. When SoTA models, including UnivFD and AIDE, are evaluated on the Chameleon dataset, they show extreme performance degradation. Interestingly, the degradation is primarily caused by the detectors classifying the fake images as real samples. These results strongly suggest that even the newer fake image detection methods only learn to detect specific fingerprints and use the real category as a sink class.

To address these limitations, we propose SimLBR, a simple framework for fake image detection using latent blending regularization. SimLBR formulates fake image detection as the task of learning a tight decision boundary around the real image distribution, producing detectors that are largely generator-agnostic and robust to future, unseen generative models. Our key idea is to introduce Latent Blending Regularization (LBR), a regularization technique that perturbs real image embeddings with controlled amounts of information from fake samples, making real-image classification non-trivial during training. Crucially, this is made possible by operating entirely in the semantically rich latent space of DINOv3, whose expressive representations preserve high-level structure and semantic information, enabling meaningful interpolation between real and fake samples. Our method is simple and efficient, requiring under three minutes for training (after precomputing embeddings) on a single NVIDIA H100 GPU. For comparison, the current state-of-the-art approach, AIDE, was trained on eight NVIDIA A100 GPUs for roughly two hours, highlighting the significant computational advantage of SimLBR.

Figure~\ref{fig:wukong_tsne} shows the substantial differences in the learned decision boundary when adding LBR. Without LBR, the detector overfits to the fake samples seen during training, causing fakes from an unseen generator (red points) to be misclassified as real. Furthermore, the detector fails to model the real image distribution seen from the clear separation between train and test real images. Adding LBR forces the model to correctly capture the structure of the real-image distribution, as seen in the compact blue/orange cluster, and accurately classifies any sample outside this region as fake. We show detailed visualizations in the supplementary. 

In addition, we argue that evaluating detectors solely by accuracy provides an incomplete picture of their real-world utility, as it reflects performance only on currently available generators. Therefore, we introduce additional reliability and robustness metrics to assess how well a model is expected to perform under varied or worst-case conditions. We summarize the main contributions of our work as the following:
\begin{itemize}
    \item We formulate fake image detection as a problem of learning tight decision boundaries around the real image distribution and using the fake category as a \textit{sink} class. 
    \item We introduce SimLBR, a simple, yet, highly effective and efficient method for AI-generated image detection using latent blending regularization.
    \item We argue for the strong need for reliability and robustness assessment of detector models, and thus, propose risk-adjusted metrics and worst-case estimates.
    \item We demonstrate that our model is both accurate and reliable, exhibiting extreme robustness, particularly when evaluated on hard-to-detect AI-generated samples.
\end{itemize}

\section{Preliminaries}
\subsection{Solving Fake Image Detection By Learning Tight Boundaries Around Real Images}
In our work, we propose to tackle fake image detection as a problem of learning tight decision boundaries around the real image distribution. Several prior works have shown that solving the fake image detection objective results in learning a tight boundary around the fake samples in the training set, causing the real category to act as a sink class~\cite{ojha2023towards}. This is a highly undesirable property since the distribution of fake images is continually evolving. Newer generative models are constantly introduced, which have completely different structures and artifacts, pushing them outside the tightly learned boundary, resulting in misclassifications.

On the contrary, the distribution of real images remains relatively constant. While there can be distribution shifts caused by improved camera sensors, these shifts are minor and often irrelevant for large periods of time. With this in mind, we propose that the ideal solution to solve fake image detection is to try and learn a tighter boundary around the real image distribution, and use the fake category as the sink class. Intuitively, this make more sense than the inverse because anything that is not real, can be placed in the fake category, irrespective of the nature and origin of the fake sample. Thus, learning to model the real distribution is more important than learning to model the fake distribution for AI-generated image detection.

While some existing methods~\cite{scholkopf2001ocsvm, ruff2018deep, hendrycks2017baseline} have attempted modeling real image distribution by framing the problem as an anomaly detection task, this strategy is insufficient for the task of fake image detection. In the extremely high-dimensional space of natural images, the assumption that photorealistic fake images will naturally fall outside the learned real-image manifold is often unrealistic. Effective discrimination therefore requires either explicit inductive biases that capture the real/fake separation or training signals derived from fake images that guide the model toward a meaningful boundary. We experimentally validate this intuition by showing that methods in anomaly detection significantly underperform in the supplemental materials. 

In our work, we propose to learn a tight decision boundary around the real image distribution by using guidance from fake images during training. To achieve this, SimLBR blends small amounts of fake image information to the real images during training, and label them as \textit{fake} category. By choosing an appropriate latent space, we can merge relevant real and fake information, using simple embedding arithmetic. Forcing the model to classify real images with even low amounts of fake information as fake leads the model to learn a tighter boundary around the uncorrupted real-image distribution, thereby allowing it to accurately classify fake images from completely unseen generators during inference. 

\subsection{Reliability in Detection}
\label{sec:reliability}
Most existing fake image detection works primarily report average accuracy as the key measure of model performance. While accuracy is a useful indicator of a detector’s capability, it provides an incomplete view of real-world reliability. When evaluated across multiple generative models, detectors often exhibit high variance in performance. For instance, a model may achieve over 99\% accuracy on one generator but drop to nearly 50\% on another. Such instability raises a fundamental question: \textbf{how much can we trust a detector’s predictions when faced with an unknown generative model?}

The true utility of a fake image detector lies not in its performance on current generative models but in its robustness to the next generation of models that remain unknown at training time. In high-stakes or safety-critical scenarios, a reliable detector must therefore demonstrate not only high average accuracy but also stable performance across diverse generative sources. Variance across generators, thus, becomes a critical dimension of evaluation, as it captures the risk and uncertainty associated with using a detector on open-world unknown generative models.

To quantify this trade-off between accuracy and stability, we introduce \textit{Reliability} as a key metric for evaluating fake image detectors. Adapting directly from the Sharpe ratio~\cite{sharpe1994ratio}, which measures excess return relative to risk, we define a detector's reliability as the ratio between a model’s excess performance and its variance across different generative models:
\begin{equation}
\text{Reliability} = \frac{\mu_{\text{acc}} - A_{\text{base}}}{\sigma_{\text{acc}}}
\label{eq:reliability}
\end{equation}
Here, $\mu_{\text{acc}}$ denotes the mean accuracy across all evaluated generative models, $\sigma_{\text{acc}}$ represents the corresponding standard deviation, and $A_{\text{base}}$ is a baseline “risk-free” accuracy from a naive model. A higher reliability score indicates a detector that achieves both high accuracy and low uncertainty, providing a principled measure of how confidently one can expect the detector to generalize to unseen generative models. 

\subsection{An Upper-Bound on Worst-Case Performance}
\label{sec:worst}
As discussed in Section~\ref{sec:reliability}, the key challenge for AI-generated image detectors lies in their ability to generalize to future generative models that exhibit strong distribution shifts relative to the training data. Therefore, it is essential to evaluate a detector’s robustness under a worst-case scenario. To approximate this setting, we define an upper bound on worst-case performance as the detector’s lowest accuracy across all evaluated generative models. This measure represents the expected performance of a detector when confronted with a future generator that deviates most significantly from the training distribution. Intuitively, a good detector should maintain a high bound of performance, since real-world generative models are likely to perform no better than this worst-case estimate.

%% file: sec/related.tex
\section{Related Works}
\subsection{Synthetic Image Generation}
With rapid advancements in AI technologies, synthetic image generation has seen significant improvements in terms of semantic quality, perception, and efficiency. Generative Adversarial Networks (GANs)~\cite{goodfellow2020generative} gained popularity and were extensively used to generate realistic and high-quality images. This success led to various applications, including CycleGAN~\cite{zhu2017unpaired}, which performs unsupervised image-to-image translation; ProGAN~\cite{karras2017progressive}, which can generate very high-resolution images; and StyleGAN~\cite{karras2019style} for style transfer between images. BigGAN~\cite{brock2018large} introduced orthogonal regularization, which significantly improved the quality of generated images. In parallel to these developments, Diffusion Models (DMs)~\cite{ho2020denoising} emerged as an alternative generative framework that outperformed GANs in terms of quality, diversity, and training efficiency. Since their introduction, numerous generative pipelines across academia and industry have incorporated DMs and have achieved state-of-the-art performance, such as ADM~\cite{dhariwal2021diffusion}, GLIDE~\cite{nichol2021glide},  SD~\cite{rombach2022high}, SD3~\cite{esser2024scaling}, DALLE-2~\cite{ramesh2022hierarchical}, MidJourney~\cite{midjourney} and Wukong~\cite{Wukong}. Moreover, DMs have been successfully utilized in various pipelines, including image editing~\cite{sushko2025realedit,fu2025univg,nano}, inpainting~\cite{kim2025rad} and controlled image generation~\cite{sun2024anycontrol,zhang2025easycontrol}.  

As these generative models continue to improve, producing images that are indistinguishable from real ones, distinguishing between authentic and synthetic content becomes significantly more challenging.

\subsection{AI-generated Image Detection}
Detecting AI-generated images has been a long-standing challenge in computer vision and media forensics. As generative models continue to improve, there is a pressing need to develop robust AI-generated image detection methods. Traditional methods used image features in the spatial domain to detect artifacts in fake images not present in real images.~\citet{wang2020cnn} introduced image augmentation to improve performance while~\citet{girish2021towards} proposed an iterative algorithm to detect fingerprints in fake images generated by GANs. Fusing~\cite{ju2022fusing} uses two-branch model to incorporate local and global image features. Since then, many works proposed to utilize various image properties such as texture~\cite{zhong2023rich}, color~\cite{he2019detection} and frequency components~\cite{frank2020leveraging, liu2022detecting}. However, majority of such methods failed to generalize well beyond the image generators they were trained on. To address this, UnivFD~\cite{ojha2023towards} was proposed which uses CLIP's latent space for fake image detection. The framework uses a database of real and fake image features to classify a query image using the nearest neighbor method. DIRE~\cite{wang2023dire} uses diffusion model's reconstruction loss while NPR~\cite{tan2024rethinking} used neighboring pixel relationships to detect fake images. Methods like PatchCraft~\cite{zhong2023patchcraft} uses pixel-wise correlation between texture features to detect fake images while AIDE~\cite{yansanity} used DCT features along with CLIP features to detect fake images. Several works in face forgery detection~\cite{nguyen2024laa, Shiohara_2022_CVPR, tian2024real} create pseudo-fakes by blending real images to prevent learning generator specific artifacts. However, since these methods operate in the pixel space, they are limited in their applicability and utility. In our work, we propose learning general AI-generated image detector using a strong regularizer that forces a model to learn generator-agnostic decision boundary.

%% file: sec/2_method.tex
\begin{figure}[!t]
\begin{center}
\includegraphics[width=\linewidth, scale=0.3]{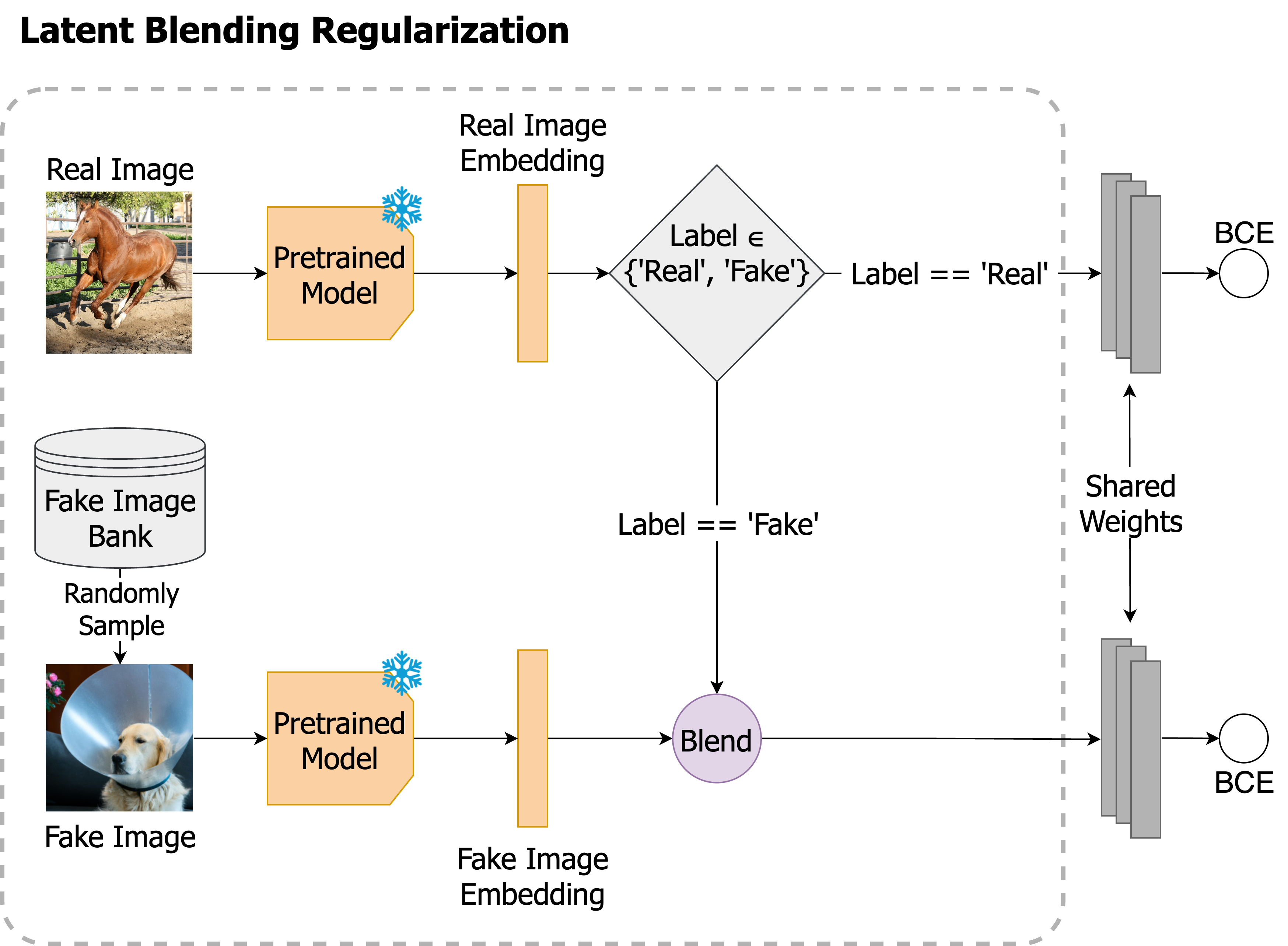}
\end{center} 
\caption{\textbf{Framework of SimLBR: } Our model samples either a real or a fake label for a real image in the training set. If a fake label is sampled, we blend small amounts of fake image information in a pretrained latent space. Solving this objective forces the model to learn a tighter decision boundary around the unperturbed real image distribution enabling better generalization to unseen generative models.}
\label{fig:framework}
\end{figure}

\section{Methodology}
In this section, we will describe in detail the motivation and design behind our framework SimLBR. 
\label{sec:method}
\subsection{Motivation}
Current AI-generated image detection models tend to learn a tight decision boundary around the fake images used during training~\cite{rajan2025staypositive}. When optimizing a binary classification objective, these models often converge to a trivial solution by exploiting low-level artifacts specific to the training fakes without ever learning meaningful discriminators between the real and fake distributions.

We argue that it is more principled to learn a tight boundary around the real image distribution, as it remains relatively unchanged over time compared to the rapidly evolving landscape of generative models. To encourage this behavior, we introduce a regularization strategy that makes the classification of real images deliberately more difficult. Instead of directly distinguishing between real and fake images, we reformulate the objective as distinguishing between real images and real images perturbed with fake information. During training, a real image $R$ is partially blended with a fake image $F$, and the resulting composite is labeled as fake. Even a small injection of fake information into a real should be classified as fake. This design forces the detector to recognize only completely unaltered, truly authentic images as real, thereby compelling it to model the underlying structure of the real distribution.

While prior works have explored image blending to increase artifact diversity, our approach is fundamentally different in motivation and formulation~\cite{Shiohara_2022_CVPR, verma2019manifold}. We perform blending in the latent space of a pretrained feature extractor rather than in the pixel domain~\cite{zhang2018mixup}. This latent space blending provides several advantages. First, it enables efficient sample construction through simple vector arithmetic, making it far more robust than pixel-space methods. Second, it allows fine-grained control over the degree of fake information injected, enabling the creation of a virtually unlimited number of perturbed-real samples. Finally, and most importantly, blending in a semantic latent space helps make the classification of real images non-trivial during training. When blending occurs in the pixel space, low-level artifacts may be directly copied from fake images, allowing the model to detect real samples trivially by focusing on these artifacts. In contrast, latent-space blending operates on higher-level representations where such low-level cues are less likely to be preserved, forcing the detector to learn meaningful semantic distinctions and effectively enforce a tighter boundary around the true real image distribution.

\begin{table*}[!t]
\resizebox{\linewidth}{!}{
\begin{tabular}{l|
>{\columncolor[HTML]{EFEFEF}}c| ccccccc|ccc}
Method     & \begin{tabular}[c]{@{}c@{}}SD v1.4\\ (Training)\end{tabular} & \multicolumn{1}{l}{Midjourney} & \multicolumn{1}{l}{SD v1.5} & \multicolumn{1}{l}{ADM} & \multicolumn{1}{l}{GLIDE} & \multicolumn{1}{l}{Wukong} & \multicolumn{1}{l}{VQDM} & \multicolumn{1}{l|}{BigGAN} & \multicolumn{1}{l}{\textit{Mean$\uparrow$}} & \multicolumn{1}{c}{\textit{Std$\downarrow$}} & \multicolumn{1}{c}{\textit{Rel.$\uparrow$}} \\
\hline
ResNet-50  & \textbf{99.90}                                                               & 54.90                           & 99.70                        & 53.50                    & 61.90                      & 98.20                       & 56.60                     & 52.00                       & 72.09                    & 22.70    & 0.97                \\
DeiT-S     & \textbf{99.90 }                                                              & 55.60                           & 99.80                        & 49.80                    & 58.10                      & 98.90                  & 56.90                     & 53.50                       & 71.56                    & 23.29     & 0.92              \\
Swin-T     & \textbf{99.90 }                                                              & 62.10                           & 99.80                        & 49.80                    & 67.60                      & 99.10                       & 62.30                     & 57.60                       & 74.78                    & 21.17   & 1.17                \\
CNNSpot    & 96.30                                                               & 52.80                           & 95.90                        & 50.10                    & 39.80                      & 78.60                       & 53.40                     & 46.80                       & 64.21                    & 22.63         & 0.62          \\
Spec       & 99.40                                                               & 52.00                           & 99.20                        & 49.70                    & 49.80                      & 94.80                       & 55.60                     & 49.80                       & 68.79                    & 24.14               & 0.77    \\
F3Net      & \textbf{99.90}                                                               & 50.10                           & \underline{99.90}               & 49.90                    & 50.00                      & \textbf{99.90}                       & 49.90                     & 49.90                       & 68.69                    & 25.85            & 0.72       \\
GramNet    & 99.20                                                               & 54.20                           & 99.10                        & 50.30                    & 54.60                      & 98.80                       & 50.80                     & 51.70                       & 68.85                    & 24.22           & 0.77       \\
DIRE       & \textbf{99.90 }                                                              & 60.20                           & 99.80                        & 50.90                    & 55.00                      & \underline{99.20}                       & 50.10                     & 50.20                       & 70.66                    & 24.22      & 0.85             \\
UnivFD     & 84.20                                                               & 73.20                           & 84.00                        & 55.20                    & 76.90                      & 75.60                       & 56.90                     & \underline{80.30}                       & 73.29                    & 11.32         & 2.05          \\
GenDet     & 96.10                                                               & \underline{89.60}                           & 96.10                        & 58.00                    & 78.40                      & 92.80                       & 66.50                     & 75.00                       & 81.56                    & 14.39         & 2.19          \\
PatchCraft & 89.50                                                               & 79.00                           & 89.30                        & 77.30                    & 78.40                      & 89.30                      & \underline{83.70}                     & 72.40                       & 82.30                     & \underline{6.56}         & \underline{4.29}           \\
AIDE       & 
\underline{99.74}                                                              & 79.38                          & 99.76                       & \underline{78.54}                   & \underline{91.82}                     & 98.65                      & 80.26                    & 66.89                      & \underline{86.88}                    & 12.33            & 2.99       \\
\hline
SimLBR     & 98.28                                                              & \textbf{91.67}                          & 98.10                       & \textbf{96.98}                   & \textbf{92.38}                     & 97.44                      & \textbf{93.50}                    & \textbf{88.00}                      & \textbf{94.54}                    & \textbf{3.74}  & \textbf{11.91}                
\end{tabular}
}

\caption{\textbf{Evaluation on GenImage Dataset: }Following protocols from prior work, we train using the Stable Diffusion 1.4 model and evaluate on other models. We outperform the SoTA models on 5 out of 7 testset models. Most importantly, we see the largest gains when evaluating on BigGAN. While other deepfake detectors rapidly deteriorate when evaluated on an out-of-distribution GAN-based model, SimLBR shows superior generalization, which is further reflected in its standard deviation and reliability scores.}
\label{table:genimage}
\end{table*}

\subsection{A Simple Framework for Fake Image Detection Using Latent Blending Regularization}
Following prior works, we train our model on fake images from a single generator $G_s$, and evaluate over a group of unseen generator models $\mathbb{U}=\{U_1, U_2, ..., U_m\}$. Let $\mathbb{R}=\{R_1, R_2, ...R_n\}$ be the set of real images and $\mathbb{F}=\{F_1, F_2, ..., F_n\}$, the set of fake images in the training set. $I$ is a pretrained image feature extractor. 

During training, a real image $R_i$ is sampled from the training set. We randomly sample a label $y_i=\{0,1\}$ for $R_i$. If $y_i=0$, we compute the features of $R_i$ as $L_i = I(R_i)$. However, if $y_i=1$, we use the Latent Blending Regularizer (LBR) to compute the features for $R_i$.
\\

\noindent\textbf{Latent Blending Regularization (LBR): }For a real image $R_i$, with label $y_i=1$, we randomly sample a fake image $F_i$ from the training set. We compute the features of both real and fake images using I:
\relax\vspace{-15pt}
\begin{center}
    \begin{align}
    L_i^{R} = I(R_i) \qquad \quad
    L_i^{F} = I(F_i)
\label{eqn:lbr_1}
\end{align}
\end{center}
The final output representation is computed as a linear interpolation between the two latent representations using $\alpha$. This training setup is summarized in Figure~\ref{fig:framework}.
\relax\vspace{-20pt}
\begin{center}
    \begin{align}
    L_i = \alpha * L_i^{R} + (1-\alpha) * L_i^{F}
\label{eqn:lbr_2}
\end{align}
\end{center}

\noindent\textbf{Sampling Distribution for $\bm{\alpha}$:}  
Choosing an appropriate sampling strategy for $\alpha$ is crucial, as it directly influences how effectively the regularizer makes real-image classification non-trivial. Low values of $\alpha$ inject a large proportion of fake information into the real sample, producing representations that are easily identified as fake due to their strong deviation from the real distribution. Conversely, extremely high values of $\alpha$ (e.g., $\alpha = 0.99$) result in blended representations that are almost indistinguishable from real samples, forcing the model to separate nearly identical embeddings into different classes, an undesirable and unstable training scenario.  

To balance these effects, we adopt the intuition that a meaningful perturbed-real sample should retain the majority of information from the original real image while incorporating only a small proportion from the fake. Accordingly, we sample $\alpha$ from a uniform distribution between $0.5$ and an upper bound $B$, i.e.,
\begin{equation}
\alpha \sim \text{Uniform}(0.5, B)
\label{eqn:alpha_sampling}
\end{equation}
This range ensures that each blended latent remains sufficiently close to the real distribution while still containing enough fake information to regularize the detector effectively. We later verify our intuition using an ablation study.

\noindent\textbf{Training Objective: }The computed features are passed through a lightweight MLP $F_\theta$. We train this MLP using a simple Binary Cross-Entropy (BCE) objective. 
\begin{align}
\mathcal{L}_{\text{BCE}} = -\frac{1}{N} &\sum_{i=1}^{N} [ y_i \log(F_{\theta}(L_i) + \nonumber \\
&(1 - y_i) \log(1 - F_{\theta}(L_i))]
\label{eqn:bce}
\end{align}
\relax\vspace{-15pt}

%% file: sec/3_results.tex
\begin{table*}[]
\resizebox{\linewidth}{!}{
\begin{tabular}{l|ccccccccccc|c}
Method & CNNSpot & FreDect & Fusing & LNP   & LGrad & UnivFD & DIRE-G & DIRE-D & PatchCraft & NPR   & AIDE  & SimLBR \\
\rowcolor[HTML]{EFEFEF} 
\hline
\begin{tabular}[l]{@{}l@{}}ProGAN\\ (Training)\end{tabular} & \textbf{100.0}   & 99.36   & \textbf{100.0}  & 99.67 & 99.83 & 99.81  & 95.19  & 52.75  & \textbf{100.0}      & 99.79 & \underline{99.99} & 99.29  \\
\hline
StyleGAN                                                    & 90.17   & 78.02   & 85.2   & 91.75 & 91.08 & 84.93  & 83.03  & 51.31  & 92.77      & \underline{97.70}  & \textbf{99.64} & 87.38  \\
BigGAN                                                      & 71.17   & 81.97   & 77.4   & 77.75 & 85.62 & 95.08  & 70.12  & 49.7   & \underline{95.80}       & 84.35 & 83.95 & \textbf{95.52}  \\
CycleGAN                                                    & 87.62   & 78.77   & 87.0   & 84.1  & 86.94 & \underline{98.33}  & 74.19  & 49.58  & 70.17      & 96.1  & \textbf{98.48} & 88.11 \\
StarGAN                                                     & 94.6    & 94.62   & 97.0   & \underline{99.92} & 99.27 & 95.75  & 95.47  & 46.72  & \textbf{99.97}      & 99.35 & 99.91 & 88.54  \\
GauGAN                                                      & 81.42   & 80.57   & 77.0   & 75.39 & 78.46 & \textbf{99.47}  & 67.79  & 51.23  & 71.58      & 82.5  & 73.25 & \underline{97.14}  \\
StyleGAN2                                                   & 86.91   & 66.19   & 83.3   & 94.64 & 85.32 & 74.96  & 75.31  & 51.72  & 89.55      & \textbf{98.38} & \underline{98.00}  & 84.57  \\
WFIR                                                        & \underline{91.65}   & 50.75   & 66.8   & 70.85 & 55.7  & 86.9   & 58.05  & 53.3   & 85.8       & 65.8  & \textbf{94.2}  & 89.50  \\
ADM                                                         & 60.39   & 63.42   & 49.0   & 84.73 & 67.15 & 66.87  & 75.78  & \textbf{ 98.25}  & 82.17      & 69.69 & \underline{93.43} & 85.89  \\
Glide      & 58.07   & 54.13   & 57.2   & 80.52 & 66.11 & 62.46  & 71.75  & \underline{92.42}  & 83.79      & 78.36 & \textbf{95.09} & 90.88  \\
Midjourney                                                  & 51.39   & 45.87   & 52.2   & 65.55 & 65.35 & 56.13  & 58.01  & \underline{89.45}  & \textbf{90.12}      & 77.85 & 77.2  & 75.54  \\
SD v1.4                                                     & 50.57   & 38.79   & 51.0   & 85.55 & 63.02 & 63.66  & 49.74  & 91.24  & \textbf{95.38}      & 78.63 & \underline{93.00}  & 85.51  \\
SD v1.5                                                     & 50.53   & 39.21   & 51.4   & 85.67 & 63.67 & 63.49  & 49.83  & 91.63  & \textbf{95.30}       & 78.89 & \underline{92.85} & 84.43   \\
VQDM                                                        & 56.46   & 77.8    & 55.1   & 74.46 & 72.99 & 85.31  & 53.68  & 91.9   & 88.91      & 78.13 & \textbf{95.16} & \underline{93.20}  \\
Wukong                     & 51.03   & 40.3    & 51.7   & 82.06 & 59.55 & 70.93  & 54.46  & 90.9   & \underline{91.07}      & 76.11 & \textbf{93.55} & 90.08  \\
DALLE2                                                      & 50.45   & 34.7    & 52.8   & 88.75 & 65.45 & 50.75  & 66.48  & 92.45  & \underline{96.60}       & 64.90  & \textbf{96.60 } & 78.80  \\
\hline
\textit{Mean}$\uparrow$                                                        & 70.78   & 64.03   & 68.38  & 83.84 & 75.34 & 78.43  & 68.68  & 71.53  & \underline{89.31}      & 82.91 & \textbf{92.77} & 88.40  \\

\textit{Std}$\downarrow$                  & 18.91   & 21.08   & 18.04  & 9.78  & 14.26 & 16.73  & 14.46  & 21.55  & 8.89  & 11.92 & \underline{7.92}  & \textbf{6.23} \\
\textit{Reliability}$\uparrow$   & 1.10  & 0.67   & 1.02  & 3.46  & 1.78 & 1.70  & 1.29  & 1.00  & 4.42  & 2.76 & \underline{5.40}  & \textbf{6.16} 
\end{tabular}
}
\caption{\textbf{Evaluation on AIGC Dataset: }Following prior work, we train using ProGAN and evaluate on 15 other generative models. SimLBR performs well across all models, ranking third-highest in average accuracy. It is the most consistent and reliable model with the lowest standard deviation and the highest reliability score. \textbf{SimLBR is also the only model to consistently perform above $75\%$, across all generative models, depicting a relatively generator-agnostic behavior.}} 
\label{table:aigc}
\end{table*}

\begin{table*}[t]
    \vspace{-2mm}
    \resizebox{\linewidth}{!}{
    \begin{tabular}{c|cccccccccc|c}
    \toprule
    \textbf{Training Dataset}              & \textbf{CNNSpot} & \textbf{FreDect} & \textbf{Fusing} & \textbf{GramNet} & \textbf{LNP}  & \textbf{UnivFD } & \textbf{DIRE} & \textbf{PatchCraft} & \textbf{NPR} & \textbf{AIDE} & \textbf{SimLBR} \\ \midrule
    \multirow{2}{*}{\textbf{ProGAN}}       & 56.94    & 55.62            & 56.98    & 58.94      & 57.11      & 57.22       & 58.19    & 53.76  & 57.29   & 58.37 & \textbf{84.33}\\
    
    & 0.08/99.67    & 13.72/87.12      & 0.01/99.79    & 4.76/99.66      & 0.09/99.97       & 3.18/97.83    & 3.25/99.48                        & 1.78/92.82  & 2.20/98.70   & 5.04/98.46  & \textbf{75.80}/90.74\\
    
    \cmidrule(l){2-12} 
    \multirow{2}{*}{\textbf{SD v1.4}}      & 60.11    & 56.86            & 57.07    & 60.95      & 55.63     & {55.62}       & 59.71    & 56.32     & 58.13    & 62.60  & \textbf{85.57}        \\
    
    & 8.86/98.63              & 1.37/98.57       & 0.00/99.96    & 17.65/93.50     & 0.57/97.01      & 74.97/41.09      & 11.86/95.67                       & 3.07/96.35  & 2.43/100.00   & 20.33/94.38 &  \textbf{92.96}/80.06  \\
    \bottomrule
    \end{tabular}
    }
    \caption{\textbf{Evaluation on \textit{hard} testset Chameleon: }We evaluate on the testset Chameleon, which is composed of specifically picked hard-to-detect samples. We report the total accuracy and fake/real accuracy for all settings. All existing models exhibit significant deterioration in accuracy and recall. SimLBR, however, maintains its performance, showing strong robustness to generator distribution shifts.}
    \label{table:chameleon}
\end{table*}

\section{Experiments and Results}
\subsection{Implementation Details}
We use DINOv3~\cite{simeoni2025dinov3} as the feature extractor in all experiments. For efficiency, we first precompute embeddings for the entire training set and train a two layer MLP classifier with ReLU activations using these features. SimLBR is trained with a batch size of 500 for a maximum of 10 epochs. The upper bound for $\alpha$ sampling distribution is set to $0.8$. We use the Adam optimizer with a learning rate of \texttt{1e-4} and weight decay of \texttt{1e-2}. A single training run requires approximately 3 minutes after embeddings are precomputed on a single H100 GPU. Precomputing the embedding takes around 50 minutes for AIGC and 22 minutes for GenImage training setup on an H100 GPU.


\subsection{Evaluation on Existing Datasets}
We evaluate SimLBR on two widely used deepfake detection benchmarks: \textbf{GenImage and AIGC}. Following the standard evaluation protocol, we train the detector using fake images generated by a single source model and test its performance across a set of unseen generative models. Along with average accuracy, we also report the standard deviation across generators and the reliability score defined in Eq.~\ref{eq:reliability}. The reliability score provides a balanced assessment of between accuracy and stability, offering a more meaningful estimate of how well a detector is expected to generalize to unseen generators. We set $A_{base}$ to $50$, as we can design a naive random predictor that achieves $50\%$ accuracy. 
\\
\noindent \textbf{Evaluation on GenImage: }We train our model on Stable Diffusion 1.4~\cite{Rombach_2022_CVPR_SD} and evaluate it across 7 different models. The results are shown in Table~\ref{table:genimage}. Our method achieves the highest average accuracy of $94.54\%$, outperforming the SoTA by $7.66\%$. More importantly, we see one of the biggest gains in BigGAN~\cite{brock2018large}, which is a GAN-based model with vastly different artifacts. These results suggest that SimLBR learns to model the real distribution and is much less impacted by changes in the distribution of the fake images during evaluation. This property is further highlighted when we evaluate the standard deviation and reliability scores across the different generator models. Our model shows the least variance, and the highest reliability score, indicating that it maintains good performance irrespective of the changes in the fake image distribution.
\\
\noindent\textbf{Evaluation on AIGC: }We train SimLBR using fake images generated by ProGAN~\cite{karras2017progressive} and evaluate its generalization on 15 unseen generative models. Table~\ref{table:aigc} shows that our method remains competitive with state-of-the-art approaches, achieving the third-highest average accuracy. More importantly, SimLBR demonstrates strong consistency across all generators, as reflected by its extremely low standard deviation and high reliability score. \textbf{Notably, SimLBR is the only model to achieve above $\bm{75\%}$ accuracy across all generative models.} 

\begin{table}
\resizebox{\linewidth}{!}{
\begin{tabular}{l|ccccc|c}
\hline
         & CNNSpot & UnivFD & DIRE  & PatchCraft  & AIDE        & SimLBR         \\ \hline
GenImage & 50.10   & 55.20  & 50.10 & \underline{72.40} & 66.89       & \textbf{88.00} \\
AIGC     & 50.45   & 50.75  & 49.74 & 70.17       & \underline{73.25} & \textbf{75.54} \\ \hline
\end{tabular}
}
\caption{\textbf{Worst Case Performance Estimates: }This table shows the lowest performance of a detector across all testset generators. This metric serves as an upper bound for the worst-case performance of a detector. SimLBR shows the best worst-case estimate across both AIGC and GenImage generators, suggesting it is more dependable when deployed in real-world settings.}
\label{table:worst}
\end{table}

\subsection{Evaluation on Curated Hard Testset}
AIDE~\cite{yansanity} recently demonstrated that common benchmarks for AI-generated image detection consist largely of randomly sampled images that do not meaningfully challenge human or algorithmic perception. To address this, AIDE introduced Chameleon, a curated dataset designed to rigorously evaluate detector robustness. Chameleon contains AI-generated images that are perceptually deceptive, verified through a human \textit{Turing Test}, span diverse semantic categories, and maintain high resolution. State-of-the-art detectors catastrophically fail on this benchmark, particularly in detecting fake images under such perceptual difficulty.

Table~\ref{table:chameleon} presents our results on Chameleon—the results are directly taken from the AIDE~\cite{yansanity} paper. In contrast to existing methods, SimLBR does not exhibit performance degradation on this hard testset. Instead, it surpasses prior state-of-the-art models by up to 25\% in accuracy and 70\% in recall. These substantial gains highlight that SimLBR avoids the common failure mode of detectors, which often overfit to specific generators and collapse when encountering new synthesis techniques. By explicitly modeling the real image distribution, SimLBR learns a highly generator-agnostic decision boundary, enabling exceptional robustness under extreme distribution shift.

\subsection{Worst-Case Performance Estimation}
We estimate the worst-case performance of fake image detectors by analyzing their minimum accuracy across all evaluated generative models. This quantity estimates an upper bound on its true worst-case performance in the open world, where future generators may be even more challenging. Table~\ref{table:worst} reports these worst-case performance bounds for state-of-the-art methods on both the GenImage and AIGC benchmarks. SimLBR achieves the highest bound under both settings, indicating that it maintains strong performance even under large distribution shifts. This suggests that SimLBR is more well-suited for deployment in the real world with continually evolving generative models.

\begin{table*}[]
\resizebox{\linewidth}{!}{
\begin{tabular}{cc|cc|cc|c|ccccc}
\multicolumn{1}{l}{}                                    & \multicolumn{1}{l|}{}                               & \multicolumn{2}{c|}{\textbf{ProGAN}}                                                                                      & \multicolumn{2}{c|}{\textbf{SD v1.4}}                                                                                     & \textbf{DiffusionSat}                                       &                                                                                 &                                                                                 &                                                                               &                                                                               &                                                                                 \\
\multicolumn{1}{l}{\multirow{-2}{*}{\textbf{Backbone}}} & \multicolumn{1}{l|}{\multirow{-2}{*}{\textbf{LBR}}} & \textbf{Chameleon}                                          & \textbf{AIGC}                                               & \textbf{Chameleon}                                          & \textbf{GenImage}                                           & \textbf{RSFake}                                             & \multirow{-2}{*}{\textbf{Mean Acc.$\uparrow$}}                                  & \multirow{-2}{*}{\textbf{Mean Rec.$\uparrow$}}                                  & \multirow{-2}{*}{\textbf{Std$\downarrow$}}                                    & \multirow{-2}{*}{\textbf{Reliability$\uparrow$}}                              & \multirow{-2}{*}{\textbf{WCE$\uparrow$}}                                        \\ \hline
DINOv2                                                  & \xmark                                              & \begin{tabular}[c]{@{}c@{}}56.12\\ 0.64/97.83\end{tabular}  & \begin{tabular}[c]{@{}c@{}}70.77\\ 44.45/97.09\end{tabular} & \begin{tabular}[c]{@{}c@{}}62.04\\ 60.73/63.78\end{tabular} & \begin{tabular}[c]{@{}c@{}}86.74\\ 77.02/95.29\end{tabular} & -                                                           & 68.92                                                                           & 45.71                                                                           & 13.31                                                                         & 1.42                                                                          & 56.12                                                                           \\ \hline
DINOv2                                                  & \chk                                                & \begin{tabular}[c]{@{}c@{}}55.77\\ 4.13/94.58\end{tabular}  & \begin{tabular}[c]{@{}c@{}}74.75\\ 57.71/91.78\end{tabular} & \begin{tabular}[c]{@{}c@{}}62.02\\ 39.23/79.14\end{tabular} & \begin{tabular}[c]{@{}c@{}}81.49\\ 70.75/90.94\end{tabular} & -                                                           & 68.51                                                                           & 42.95                                                                           & 11.71                                                                         & 1.57                                                                          & 55.77                                                                           \\ \hline
DINOv3                                                  & \xmark                                              & \begin{tabular}[c]{@{}c@{}}59.65\\ 7.24/99.03\end{tabular}  & \begin{tabular}[c]{@{}c@{}}76.70\\ 54.07/99.33\end{tabular} & \begin{tabular}[c]{@{}c@{}}87.79\\ 89.03/86.86\end{tabular} & \begin{tabular}[c]{@{}c@{}}92.19\\ 85.80/97.81\end{tabular} & \begin{tabular}[c]{@{}c@{}}73.90\\ 48.28/99.52\end{tabular} & 78.05                                                                           & 58.88                                                                           & 12.77                                                                         & 2.19                                                                          & 59.65                                                                           \\ \hline
\rowcolor{gray!20} 
DINOv3                                                  & \chk                                                & \begin{tabular}[c]{@{}c@{}}84.33\\ 75.80/90.74\end{tabular} & \begin{tabular}[c]{@{}c@{}}88.40\\ 82.07/93.97\end{tabular} & \begin{tabular}[c]{@{}c@{}}85.57\\ 92.96/80.06\end{tabular} & \begin{tabular}[c]{@{}c@{}}94.54\\ 93.89/95.05\end{tabular} & \begin{tabular}[c]{@{}c@{}}88.88\\ 78.87/99.29\end{tabular} & \begin{tabular}[c]{@{}c@{}}88.26\\ (\textcolor{darkgreen}{+10.21})\end{tabular} & \begin{tabular}[c]{@{}c@{}}84.72\\ (\textcolor{darkgreen}{+25.84})\end{tabular} & \begin{tabular}[c]{@{}c@{}}3.94\\ (\textcolor{darkgreen}{-8.83})\end{tabular} & \begin{tabular}[c]{@{}c@{}}9.70\\ (\textcolor{darkgreen}{+7.51})\end{tabular} & \begin{tabular}[c]{@{}c@{}}84.33\\ (\textcolor{darkgreen}{+24.68})\end{tabular} \\ \hline
\end{tabular}
}
\caption{\textbf{Ablation of SimLBR: }We ablate the impact of latent blending regularization and the latent space used for blending. For each setting, we report the total accuracy and fake/real accuracy. LBR significantly improves the accuracy and recall performance in the DINOv3 space. Furthermore, it achieves much better reliability scores and worst-case estimates, resulting in more reliable and robust detectors. However, the choice of latent space is an important consideration in SimLBR as DINOv2 does not achieve the same performance boost.}
\label{table:ablate}
\end{table*}

\begin{figure}[!t] 
  \centering
  \begin{subfigure}[t]{0.48\linewidth}
    \centering
    \includegraphics[width=\linewidth]{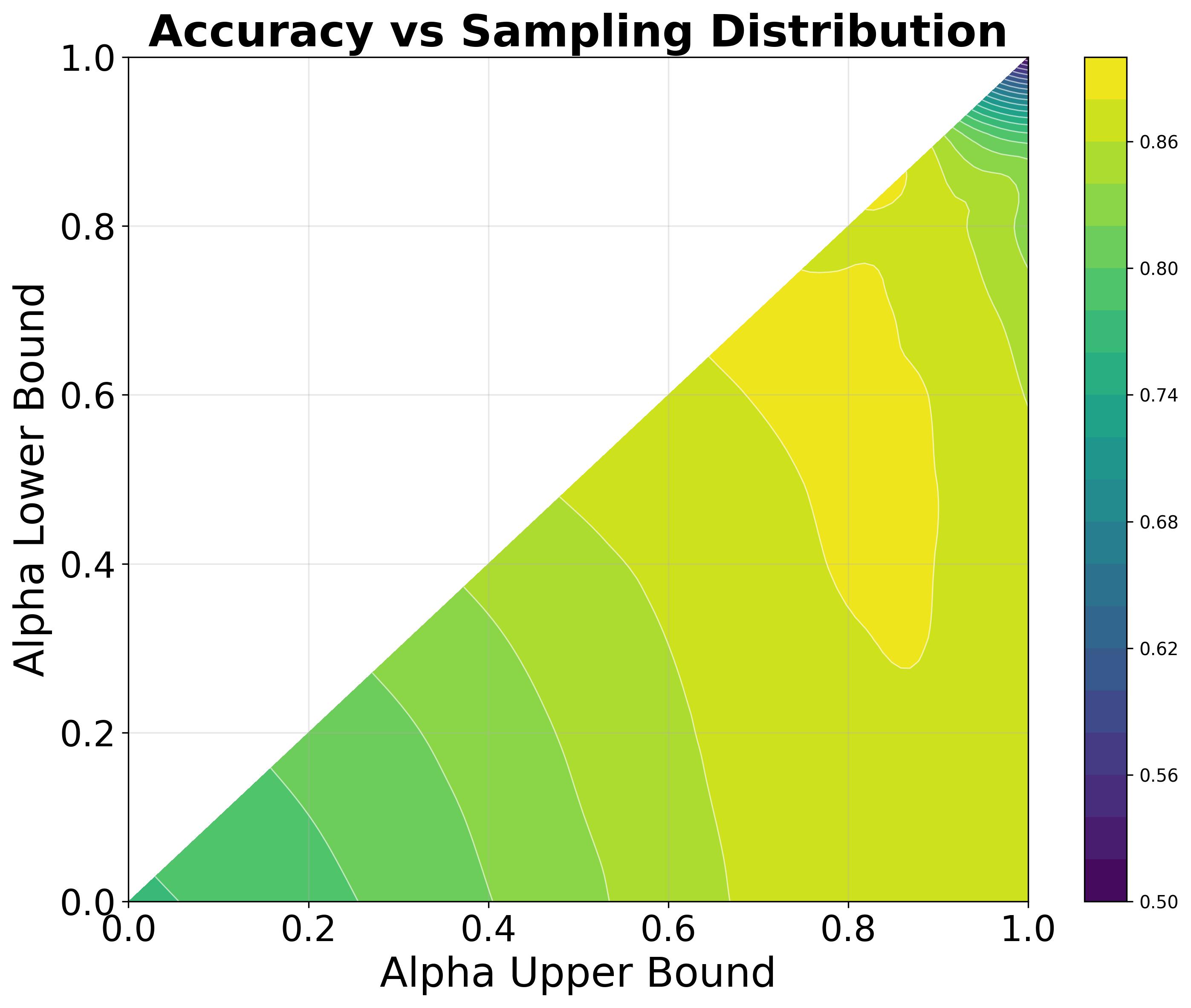}
     \subcaption{}
    \label{fig:alpha_ablate}
  \end{subfigure}\hfill
  \begin{subfigure}[t]{0.48\linewidth}
    \centering
    \includegraphics[width=\linewidth]{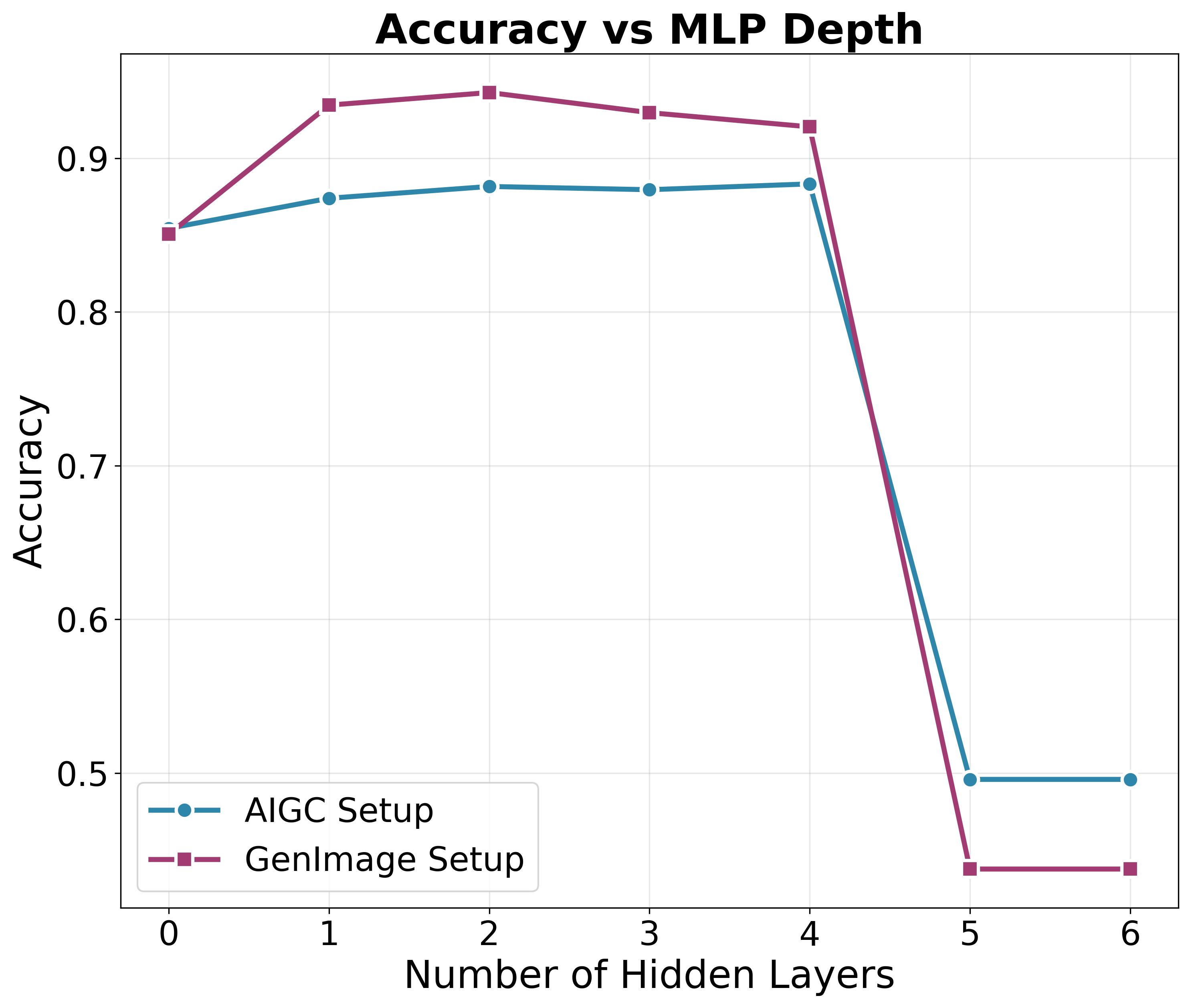}
     \subcaption{}
    \label{fig:mlp_ablate}
  \end{subfigure}
  \caption {\textbf{(a) Ablation of $\alpha$:} We train SimLBR using ProGAN and different sampling distributions for $\alpha$. Always retaining at least half the information from the real image is one of the optimal and robust choices for sampling $\alpha$. \textbf{(b) Ablation of MLP: }We try different sizes of MLP. SimLBR performs better with smaller MLPs and overfits with large number of hidden layers.}
  \label{fig:ablate}
\end{figure}

\subsection{Ablation Studies}
\textbf{Impact of LBR and Latent Space: }To isolate the contribution of latent blending regularization, we train a baseline MLP model under an identical setup without using LBR. To understand the impact of the backbone latent space, we train models using features from the older DINOv2~\cite{oquab2024dinov2}. In addition to evaluating on Chameleon, AIGC, and GenImage, we also include the RSFake dataset~\cite{tan2025rsfake1m}, a large-scale benchmark for detecting generated remote-sensing forgeries. This dataset allows us to evaluate our model under large distribution shifts. Details of the RSFake data and training protocols are provided in the supplementary material.

Table~\ref{table:ablate} summarizes the results. Adding latent blending regularization in the DINOv3 space yields substantial improvements across the datasets, achieving an average accuracy gain of 10.11\%. The effect is even more pronounced when analyzing the fake image accuracy. Specifically, fake image accuracy improves from 7.24\% → 75.80\% on Chameleon-ProGAN, 54.07\% → 82.07\% on AIGC, 89.03\%→92.96\% on Chameleon-SD, 85.80\% → 93.89\% on GenImage, and 48.28\% → 78.87\% on RSFake, corresponding to an average increase of over \textbf{25 percentage points}. 
Although the improvements on the SD v1.4–trained model appear less dramatic, we show in the supplementary that this is because the baseline detector overfits to diffusion-specific artifacts present in the training data and fails sharply when evaluated on fake images produced by GAN-based architectures with distinct generative priors. Furthermore, latent blending substantially improves the stability and reliability of the detector, as reflected by its low standard deviation, and high reliability scores / worst-case estimates. These results clearly demonstrate that LBR is essential to training reliable models that perform well in all settings, ideal for real-world deployment.

Interestingly, we do not see the same level of performance improvements when working in the DINOv2 space. The LBR method operates entirely in the latent space by injecting low amounts of fake image information into the real image latents. This setting assumes a geometrically smooth manifold, where such an interpolation operation yields a smooth change lying on the valid embedding manifold. We leave the exploration of embedding space properties that make LBR effective to future work. We hypothesize that this behavior is related to limitations in the manifold structure of weaker embedding models~\cite{kang2025clip, li2025lost, liu2024assessing}.

\noindent\textbf{Impact of Sampling Distribution for Alpha: }We perform a sensitivity analysis to examine how different sampling strategies for $\alpha$ affect model performance, as shown in Figure~\ref{fig:alpha_ablate}. Consistent with our hypothesis, sampling only low values of $\alpha$ leads to poor performance, since injecting excessive fake information makes the classification task trivial. Conversely, sampling only high values of $\alpha$ causes sharp performance degradation, as the perturbed samples become nearly identical to the unperturbed real images, making the regularization too strong. Sampling from a uniform distribution, $\text{Uniform}(0.5, B)$, strikes a balance between these extremes by generating semantically meaningful perturbed-real samples that effectively regularize the detector to model the real image distribution. The method is also robust to the choice of $B$, with values in the range $0.7 < B < 0.9$ consistently yielding optimal performance. 

\vspace{2pt}

\noindent\textbf{Impact of MLP Depth:} We train SimLBR with different MLP sizes, as shown in Figure~\ref{fig:mlp_ablate}. Our method performs well with 0-4 deep layers. However, we see rapid performance deterioration when beyond 4 layers. With larger MLPs, the model overfits excessively to the training data.

%% file: sec/conclusion.tex
\vspace{-2pt} \relax
\section{Conclusion}
In this work, we revisited the fundamental objective of AI-generated image detection and argued that reliable generalization requires learning learning a tight boundary around the real image distribution. Building on this insight, we introduced SimLBR, a simple yet highly effective framework for fake image detection that leverages Latent Blending Regularization within the semantically rich latent space of DINOv3. SimLBR achieves strong generalization across diverse generative models while being orders of magnitude more efficient than existing approaches. We also highlighted the limitations of traditional evaluation practices and introduced a Sharpe ratio–based Reliability Score and Worst-Case Estimates, which more faithfully characterize detector robustness under generator distribution shifts. 

While SimLBR demonstrates strong cross-generator generalization, it operates under the assumption that the real image distribution remains relatively stable across domains. We leave to future work to investigate the efficacy of SimLBR under conditions where these assumptions might not hold. Together, our work advances the field toward developing detectors that are not only accurate but also robust and resilient to the rapid evolution of generative models.


\section*{Acknowledgements}
We acknowledge that this manuscript has been authored by UT-Battelle, LLC under Contract No. DE-AC05-00OR22725 with the U.S. Department of Energy. The United States Government retains and the publisher, by accepting the article for publication, acknowledges that the United States Government retains a non-exclusive, paid-up, irrevocable, world-wide license to publish or reproduce the published form of this manuscript, or allow others to do so, for United States Government purposes. DOE will provide public access to these results of federally sponsored research in accordance with the DOE Public Access Plan (http://energy.gov/downloads/doe-public-access-plan).

This research used the TGI RAILs advanced compute and data resource which is supported by the National Science Foundation (award OAC-2232860) and the Taylor Geospatial Institute.

%% file: sec/X_suppl.tex
\clearpage

\onecolumn

\maketitlesupplementary

\begin{table}[t]
\begin{center}
\begin{tabular}{|c|cc|cc|}
\hline
                           & \multicolumn{2}{c|}{\textbf{GenImage}} & \multicolumn{2}{c|}{\textbf{AIGC}} \\ \cline{2-5} 
                           & ID             & OD            & ID           & OD           \\ \hline
w/o LBR                    & 94.37              & 77.70             & 85.58            & 73.26           \\
w/ LBR                     & 95.47              & 88.00             & 91.50            & 85.98           \\ \hline
\multicolumn{1}{|l|}{Gain} & 1.10               & \textbf{10.30}             & 5.92             & \textbf{12.72 }          \\ \hline
\end{tabular}
\caption{\textbf{Average accuracy on ID vs. OD testsets: }We ablate the impact of using Latent Blending Regularization on In-Distribution (ID) and Out-Distribution (OD) generative testsets. ID testset contains generators that share architectural similarities with the training generator, whereas OD testset contains generators with substantially different architectures. We observe the most significant gains from LBR in the OD testset, reasserting that LBR encourages the model to learn a generator-agnostic decision boundary.}
\label{table:supp_ablate}
\end{center}
\end{table}

\section{Evaluating LBR on Out of Distribution Generators}
In Table~\ref{table:ablate}, we presented the gains in average accuracy obtained by adding Latent Blending Regularization (LBR). Prior works~\cite{zhu2023genimage, yansanity} have shown that detectors generally perform better on generators that share architectural similarities with the training generator, as they can share similar artifacts in their generated data. Thus, to further isolate the contribution of LBR, we partition the test-time generators into two groups: In-Distribution (ID) and Out-Distribution (OD). 

The ID set contains generators whose architectural families match that of the training generator, while the OD set contains generators with fundamentally different architectures. For the AIGC benchmark, all GAN-based generators are treated as ID and diffusion-based generators as OD; for GenImage, this assignment is reversed. This partition allows us to assess whether LBR improves performance primarily on generators that are artifact-similar (ID) or on generators that introduce entirely different characteristics (OD).

Table~\ref{table:supp_ablate} shows the average accuracy across different settings. In both GenImage and AIGC benchmarks, LBR yields modest gains on the ID set, where the baseline already performs well. However, the improvements on the OD set are substantially larger, with gains of 10\% on GenImage and 12.72\% on AIGC. These results indicate that while baseline detectors overfit to artifacts present in the training generator, LBR helps the model learn a more principled separation between the real and fake distributions. As a result, SimLBR achieves significantly stronger performance on OD generators, demonstrating its ability to generalize beyond artifact-level correlations and to adapt to previously unseen generative architectures.

\section{Robustness to Image Perturbations}
Prior works~\cite{yansanity, zhong2023patchcraft} show that detectors exhibit severe performance degradation when perturbations such as JPEG Compression and Gaussian Blur are applied to the input images. We evaluate the performance of SimLBR under different compression and blur settings on the AIDE benchmark. The reported accuracy is averaged across 16 different generative models. The results are summarized in Table~\ref{table:perturbation}. Compared to the state-of-the-art models like AIDE~\cite{yansanity} and PatchCraft~\cite{zhong2023patchcraft}, SimLBR shows extreme robustness to these perturbations. SimLBR achieves the highest accuracy across all perturbations, exhibiting incredible robustness to adversarial inputs. Furthermore, unlike most prior methods~\cite{yansanity, zhong2023patchcraft, wang2020cnn}, we did not include JPEG compression or Gaussian blur augmentations during training, which indicates that SimLBR is truly robust to unseen perturbations.

\begin{table*}
\begin{tabular}{lc|cccc|cccc}
    \toprule
    \multirow{2}{*}{\textbf{Method}} & \multirow{2}{*}{\textbf{Original}}                      & \multicolumn{4}{c|}{\textbf{JPEG Compression}}  & \multicolumn{4}{c}{\textbf{Gaussian Blur}}        \\
                                                 &       & \textbf{QF=95}        & \textbf{QF=90}    & \textbf{QF=75} & \textbf{QF=50}    & \bm{$\sigma = 1.0$} & \bm{$\sigma = 2.0$} & \bm{$\sigma = 3.0$}& \bm{$\sigma = 4.0$} \\ \midrule
    \textbf{CNNSpot} &70.78       & 64.03                 & 62.26      & 60.65 & 59.66            & 68.39                   & 67.26  & 67.13 & 65.85                 \\
    \textbf{FreDect} &64.03        & 66.95                 & 67.45     & 66.64&     65.33         & 65.75                   & 66.48  &  68.58 &69.64                \\
    \textbf{Fusing} &68.38         & 62.43                 & 61.39     & 59.34 &      57.41        & 68.09                   & 66.69    &66.02 & 65.58                 \\
    \textbf{LNP}  &83.84           & 53.58                 & 54.09    & 53.02 &  52.85             & 67.91                   & 66.42 & 66.2 & 62.69                    \\
    \textbf{LGrad}  &75.34         & 51.55                 & 51.39   & 50.00  &     50.00       & 71.73                   & 69.12    & 68.43 & 66.22                 \\
    \textbf{DIRE-G} &68.68         & 66.49                 & 66.12    & 65.28 &      64.34        & 64.00                   & 63.09   & 62.21 & 61.91                  \\
    \textbf{UnivFD} &78.43         & {74.10}     & {74.02}  & {69.92} & {68.68}   & 70.31                   & 68.29   &64.62 & 61.18                  \\
    \textbf{PatchCraft} &89.31     & 72.48                 & 71.41       & 69.43 &   67.78         & {75.99}       & {74.90}    & {73.53} & {72.28}                 \\ 
    \textbf{AIDE} &\textbf{92.77 }          & 75.54        & 74.21   & 70.64	& 69.60
          & 81.88          & 80.35   & 80.05 &	79.86 \\ \midrule 
    \textbf{SimLBR} & 88.40 & \textbf{ 86.04}  &  \textbf{83.92}  & \textbf{78.98}	& \textbf{73.48}  &  \textbf{88.03} & \textbf{86.65}  & \textbf{86.39} & \textbf{86.30}\\      \bottomrule
    \end{tabular}
 
    \caption{\textbf{Robustness to JPEG Compression and Gaussian Blur: }Following prior work, we evaluate the robustness of our model when different amounts of JPEG Compression and Gaussian Blur are applied to the input images. SimLBR is highly robust, showing minimal degradation even under strong image perturbations.}
    \label{table:perturbation}
\end{table*}

\begin{figure*}[!t]
  \centering
  {\footnotesize\textbf{w/o Latent Blending Regularization}}\\[2pt]
    \begin{minipage}[t]{0.19\linewidth}
    \includegraphics[width=\linewidth]{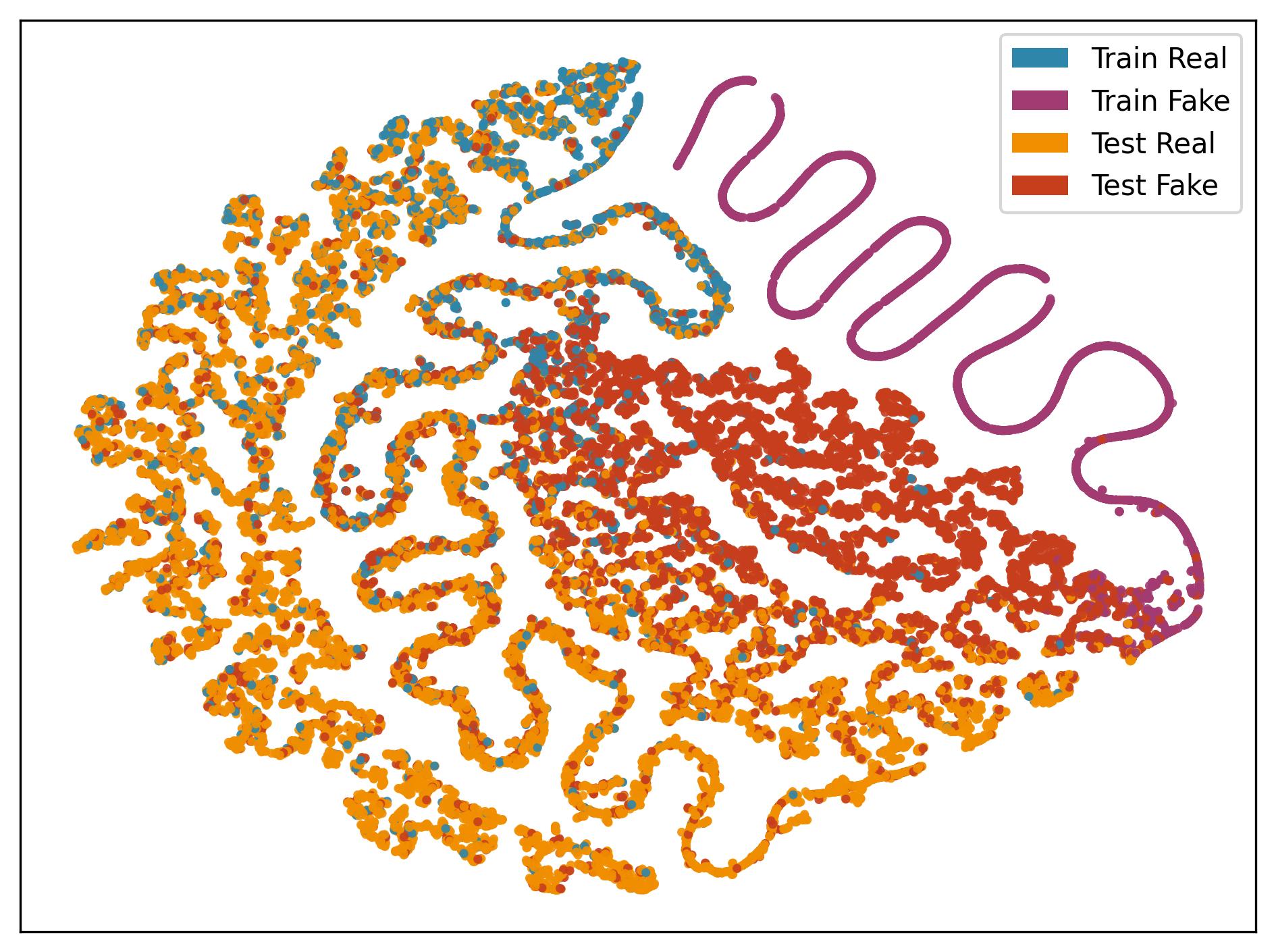}
  \end{minipage}\hfill
  \begin{minipage}[t]{0.19\linewidth}
  \includegraphics[width=\linewidth]{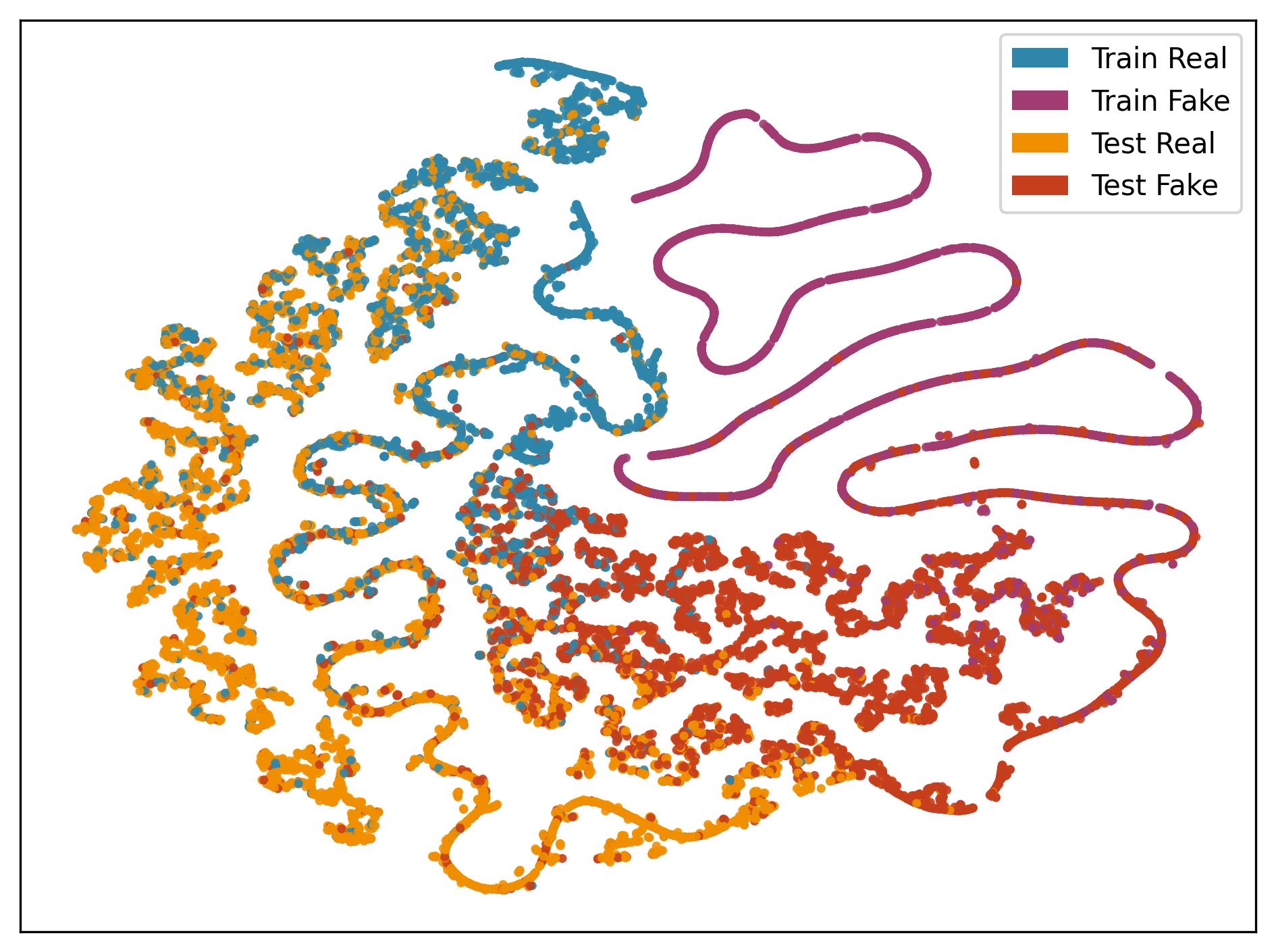}
  \end{minipage}\hfill
  \begin{minipage}[t]{0.19\linewidth}
    \includegraphics[width=\linewidth]{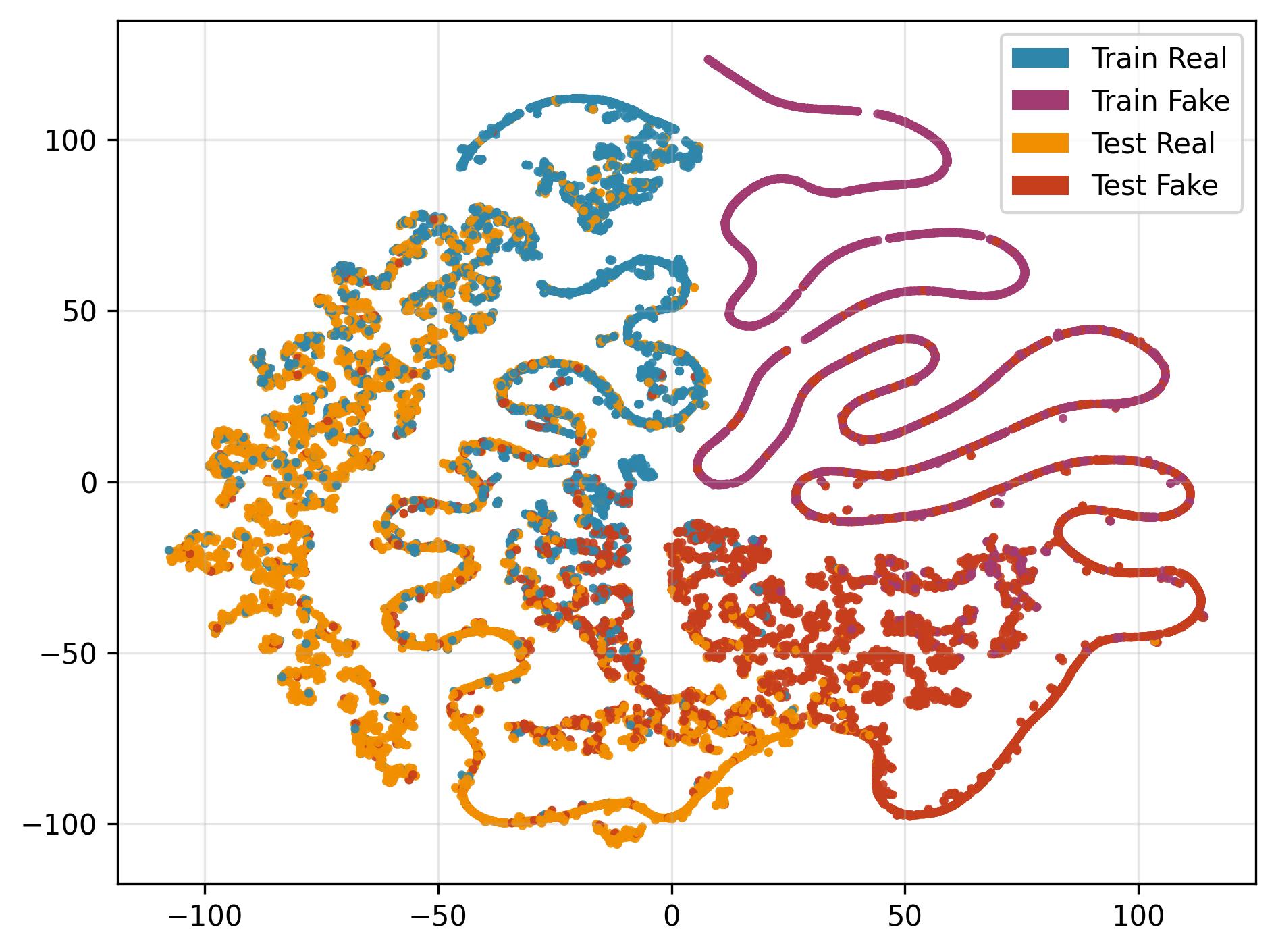}
  \end{minipage}\hfill
  \begin{minipage}[t]{0.19\linewidth}
    \includegraphics[width=\linewidth]{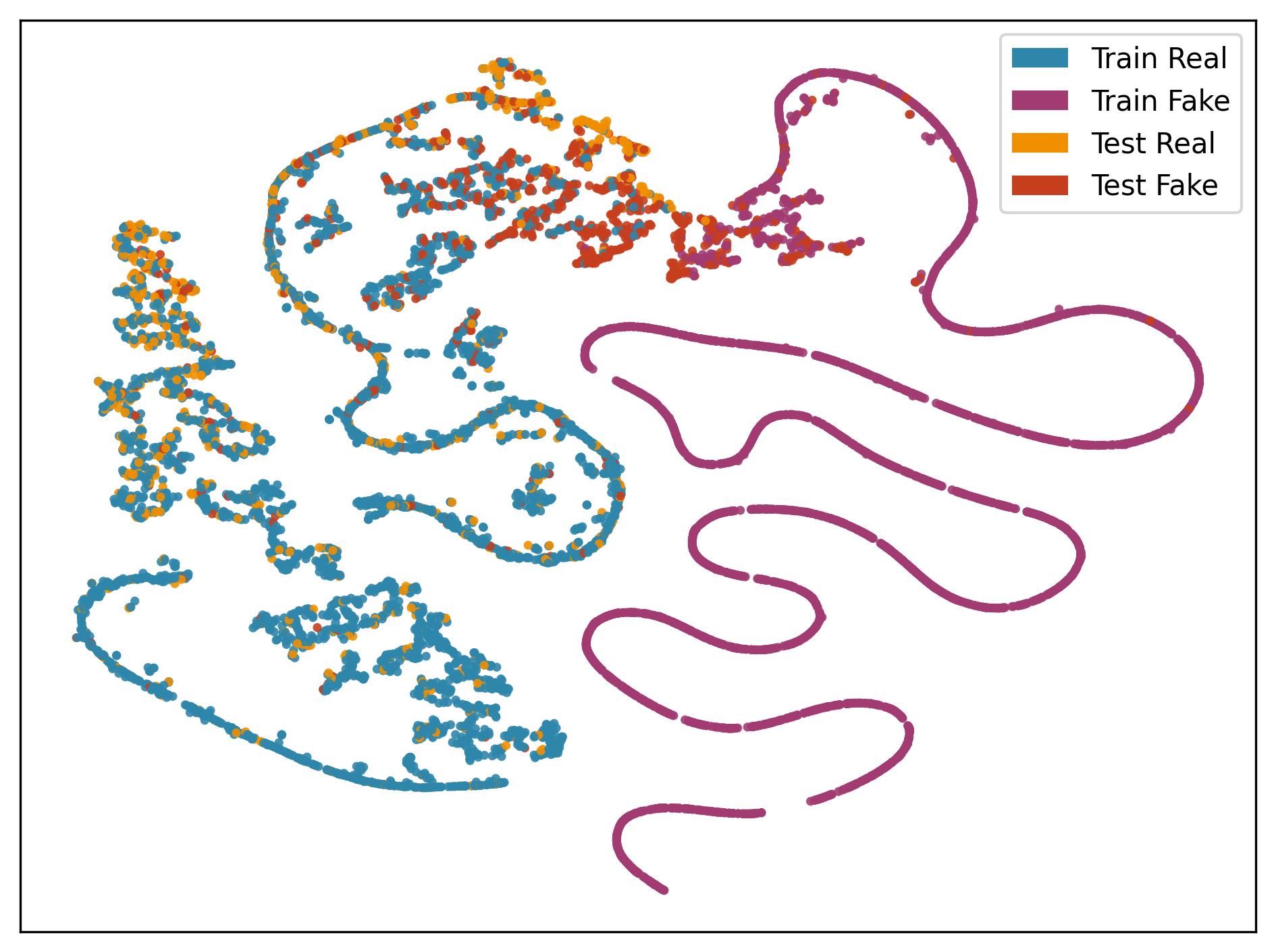}
  \end{minipage}\hfill
  \begin{minipage}[t]{0.19\linewidth}
     \includegraphics[width=\linewidth]{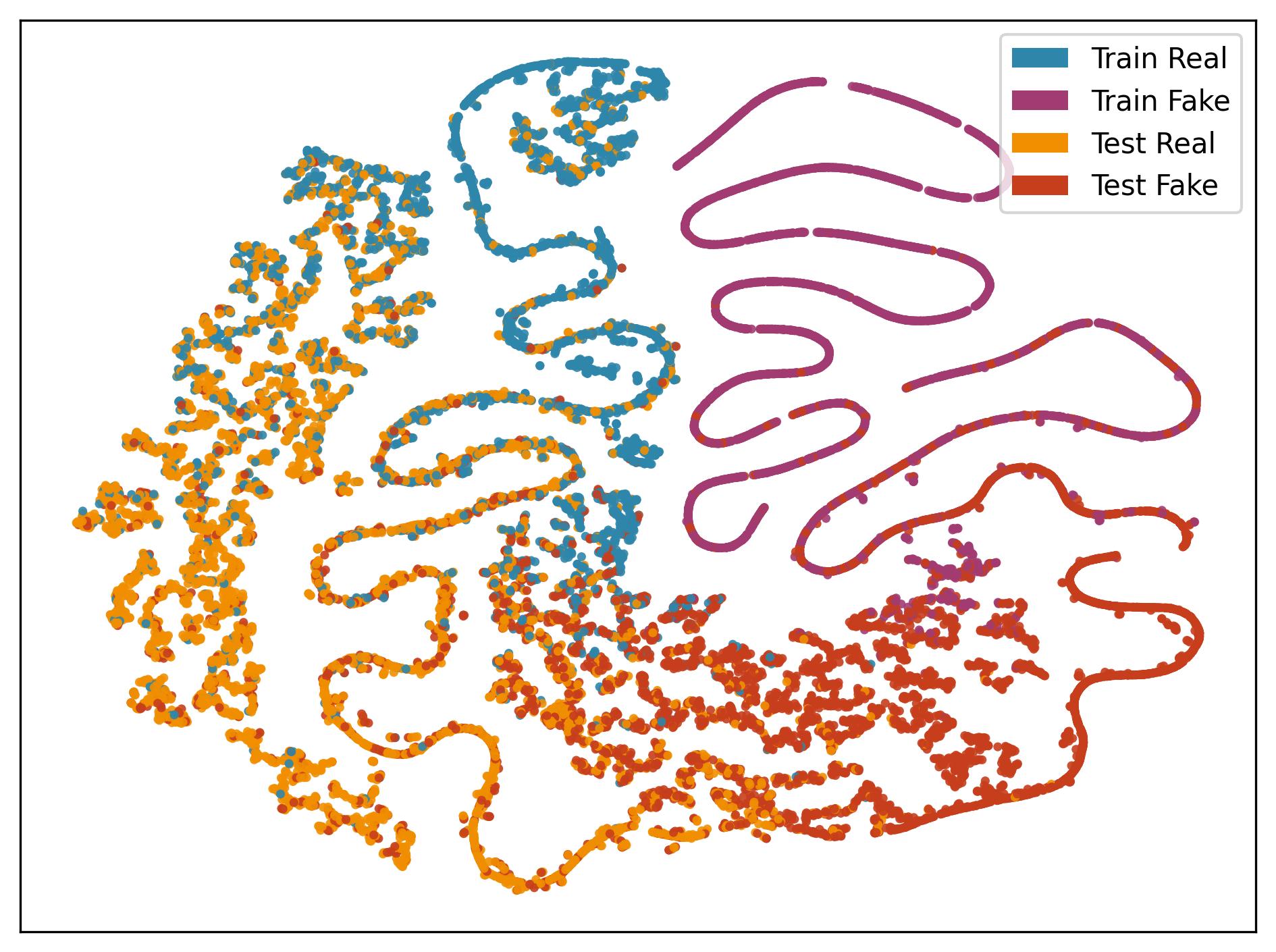}
  \end{minipage}\\[4pt]
  
  \begin{minipage}[t]{0.19\linewidth}
    \includegraphics[width=\linewidth]{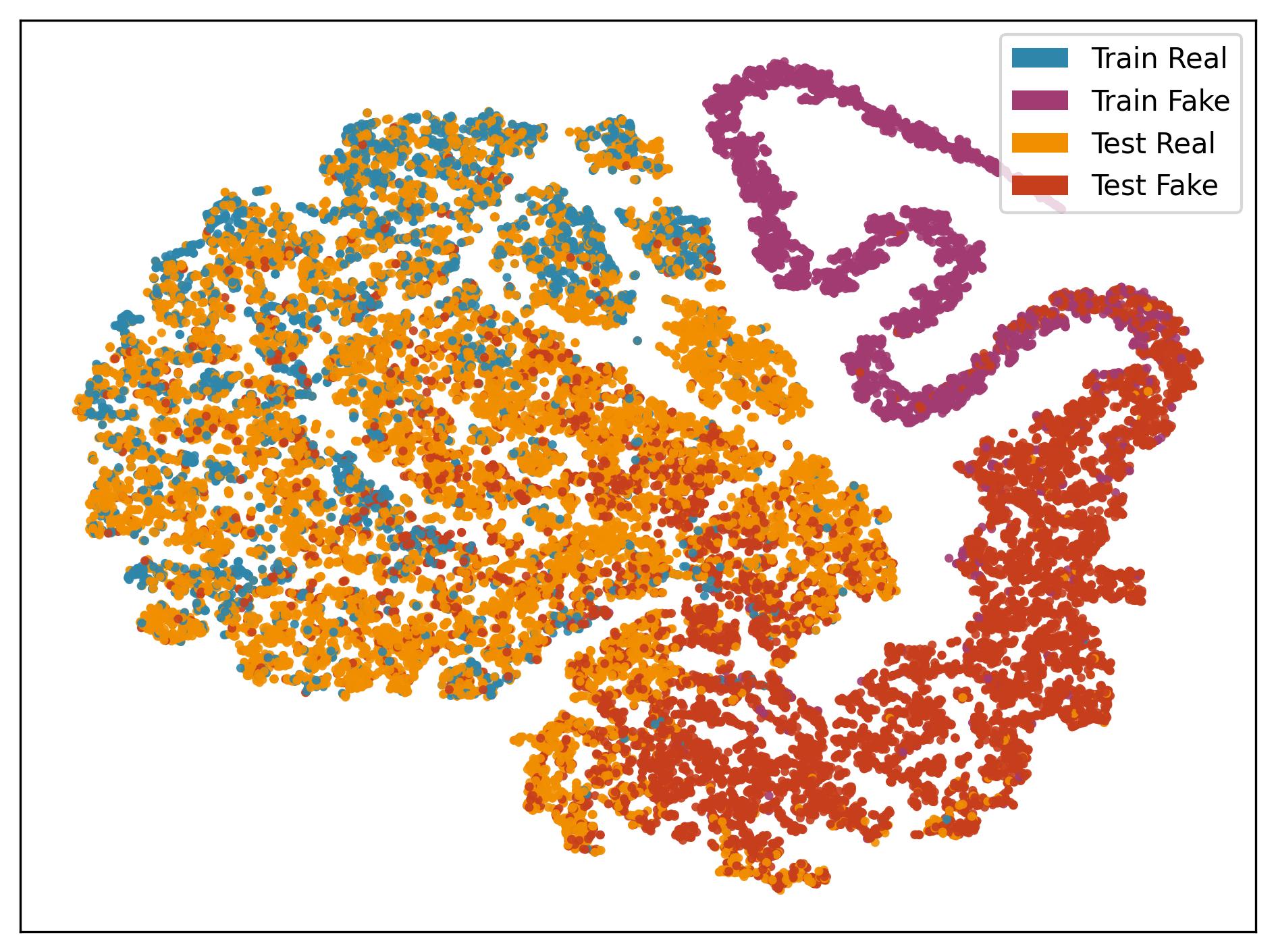}
    \subcaption[]{Chameleon}
  \end{minipage}\hfill
    \begin{minipage}[t]{0.19\linewidth}
    \includegraphics[width=\linewidth]{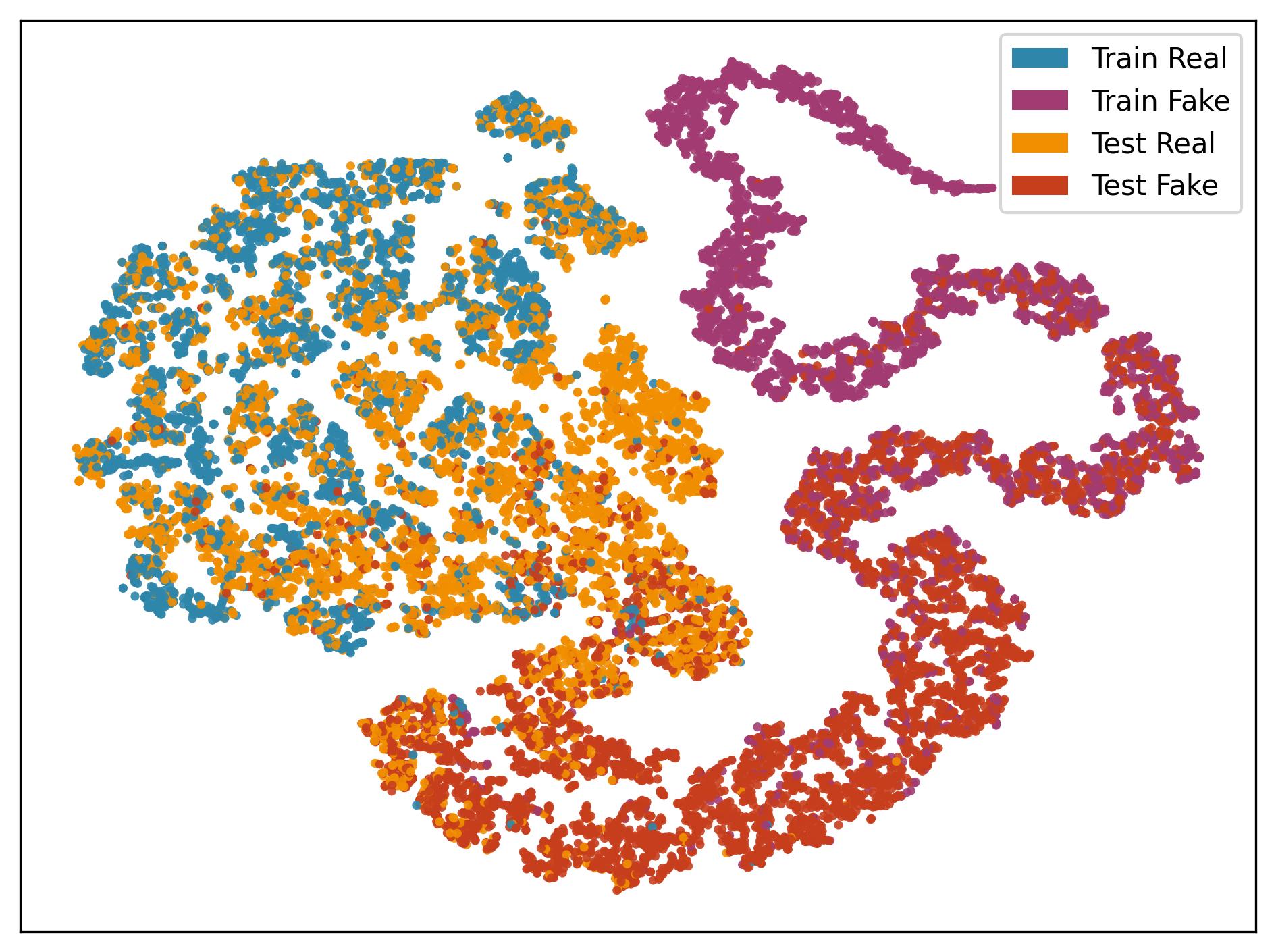}
    \subcaption[]{Wukong}
  \end{minipage}\hfill
  \begin{minipage}[t]{0.19\linewidth}
    \includegraphics[width=\linewidth]{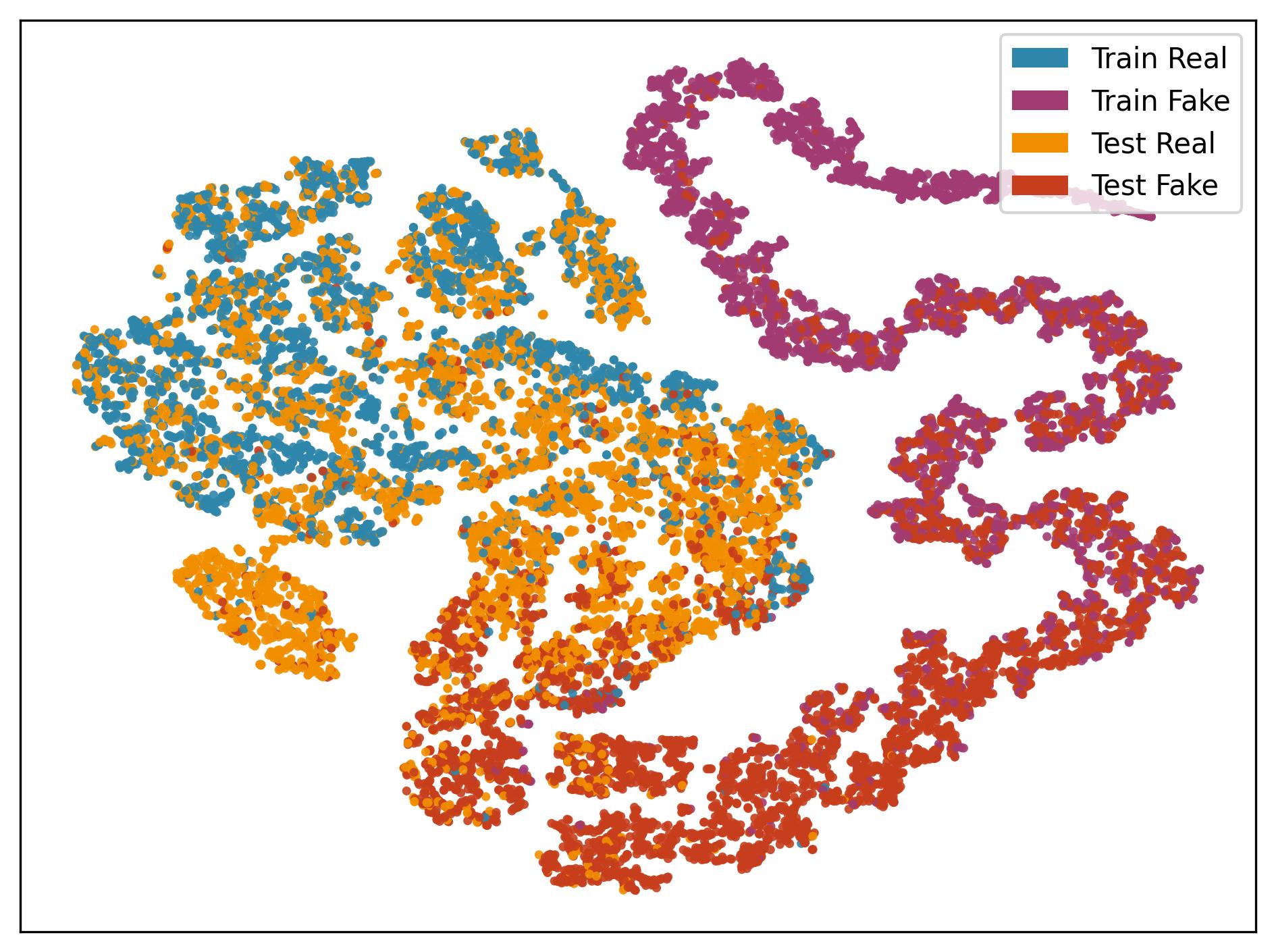}
    \subcaption[]{Glide}
  \end{minipage}\hfill
  \begin{minipage}[t]{0.19\linewidth}
    \includegraphics[width=\linewidth]{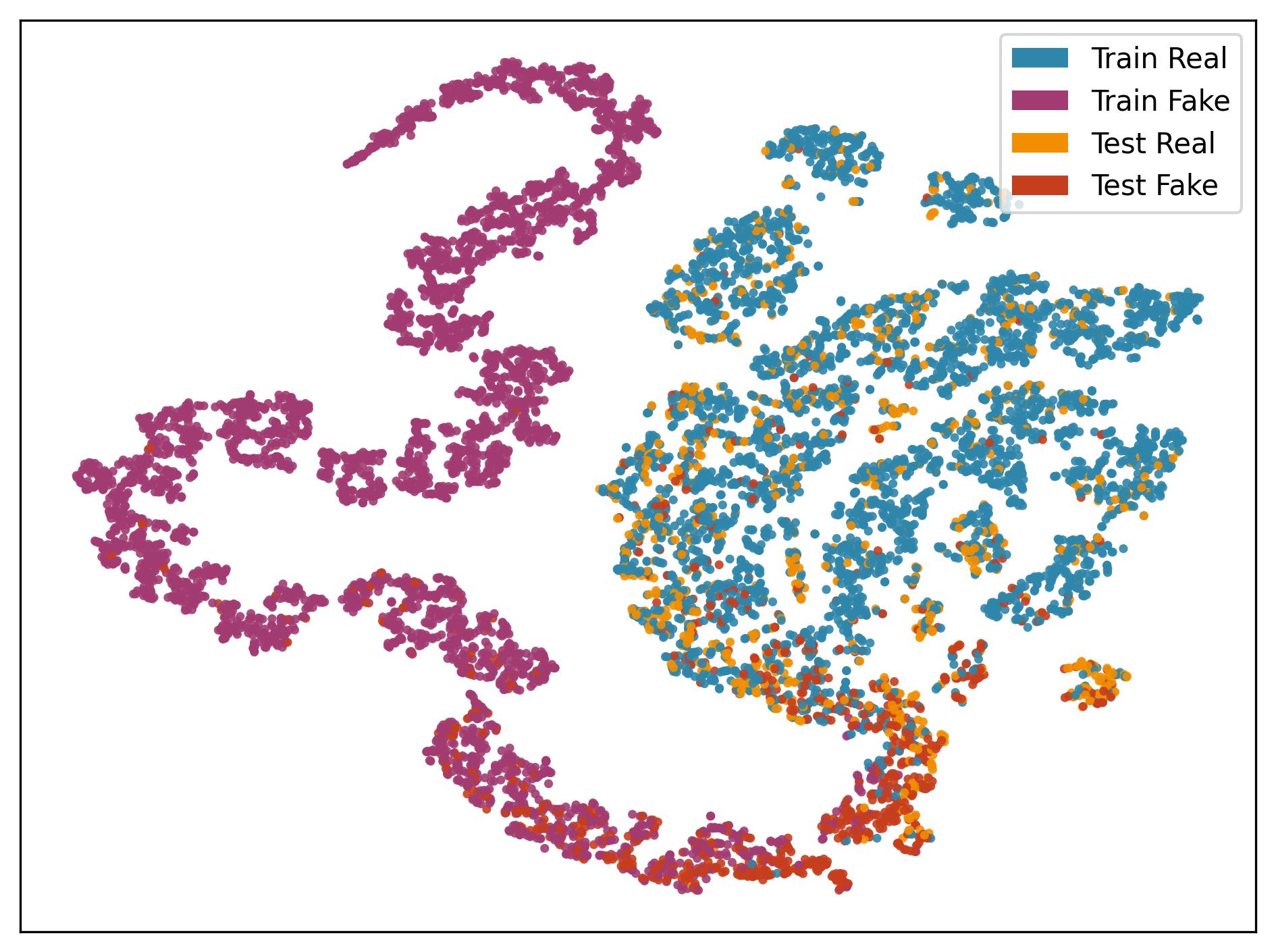}
    \subcaption[]{DALLE2}
  \end{minipage}\hfill
  \begin{minipage}[t]{0.19\linewidth}
    \includegraphics[width=\linewidth]{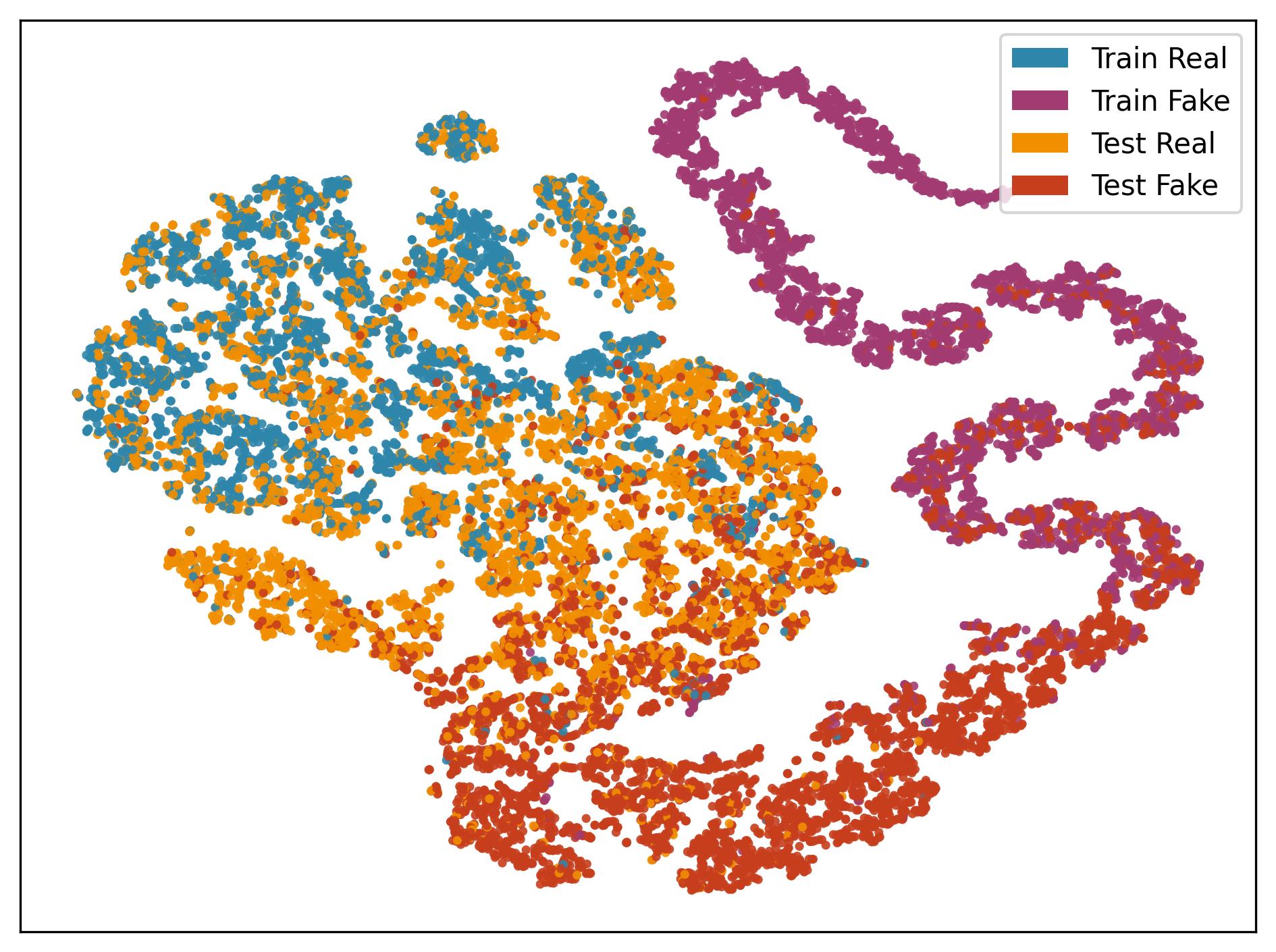}
    \subcaption[]{ADM}
  \end{minipage}\\
  {\footnotesize\textbf{w/ Latent Blending Regularization}}\\[2pt]

  \captionof{figure}{\textbf{t-SNE plot visualization: }Figure shows t-SNE plot from the penultimate layer of models trained with and without our approach. While the existing approach overfits to fake samples in the training set, our method correctly models the separation between real and fake image distributions, allowing generalization to any generator.}
  \label{fig:grid_2x5}

    \label{img:t-SNE full}
    \end{figure*}

\section{Visualization of SimLBR Embeddings}
\label{sec:t-SNE}
Figure~\ref{img:t-SNE full} shows t-SNE visualizations of embeddings from final deep layer of the MLP trained with and without Latent Blending Regularization. For the training set, we use images generated by the ProGAN model and visualize embeddings across five datasets with diverse generated images. We see that naively training the model to differentiate between real and fake images results in models that overfit to the training data and do not preserve meaningful structures between real and fake distributions. LBR forces the model to learn a more meaningful separation between the real and generated image distribution by learning a tight boundary around the real images, which is visible as the compact \textcolor{tblue}{blue}/\textcolor{orange}{orange} cluster. These visualizations strongly suggest that LBR appropriately regularizes the training objective, forcing the model to learn generator-agnostic decision boundaries.

\section{RSFake-1M Dataset}
\label{sec:rsfake}
In addition to evaluating on natural-image benchmarks such as AIGC and GenImage, we also conduct experiments on RSFake-1M~\cite{tan2025rsfake}, a large-scale dataset designed for detecting diffusion-generated satellite imagery forgeries. The objective of including this dataset is to evaluate whether LBR improves performance across highly varied image distributions. RSFake-1M contains 500,000 synthetic satellite images generated from 10 satellite-specific diffusion models, paired with 500,000 authentic satellite images. The synthetic generators include: DiffusionSat-512~\cite{khanna2024diffusionsat}, DiffusionSat-256~\cite{khanna2024diffusionsat}, GeoRSSD~\cite{zhang2024rs5m}, SDFRS~\cite{yuan2023efficient}, GeoSynth-Text~\cite{sastry2024geosynth}, GeoSynth-Sam~\cite{sastry2024geosynth}, GeoSynth-Canny~\cite{sastry2024geosynth}, CRSDiff~\cite{tang2024crs}, MapSat~\cite{espinosa2023generate}, and RSPaint~\cite{immanuel2025tackling}.

For our experiments, we train SimDLR using DiffusionSat-512 and DiffusionSat-256, and evaluate on held-out test sets from all ten models to assess cross-generator generalization. As the detection backbone, we use the DINOv3-Satellite model, pretrained on 493M satellite images. Since a satellite-specific version of DINOv2 is not available, the RSFake-1M evaluations are performed using only the DINOv3 backbone.


\section{Comparison to Anomaly Detection}
\begin{figure}[!t] 
  \centering
  \begin{subfigure}[t]{0.48\linewidth}
    \centering
    \includegraphics[width=\linewidth]{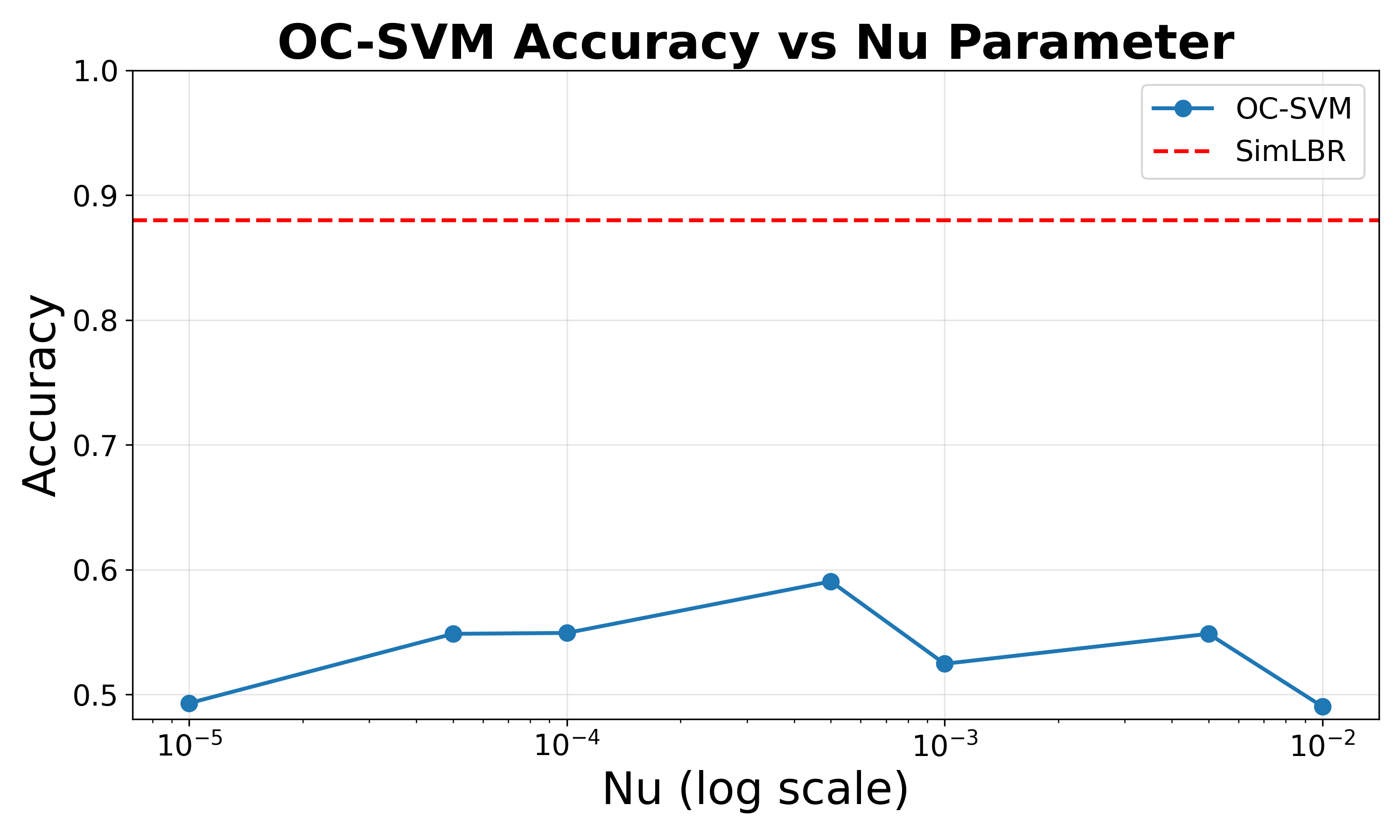}
     \subcaption{AIGC}
    \label{fig:ocsvm_aigc}
  \end{subfigure}\hfill
  \begin{subfigure}[t]{0.48\linewidth}
    \centering
    \includegraphics[width=\linewidth]{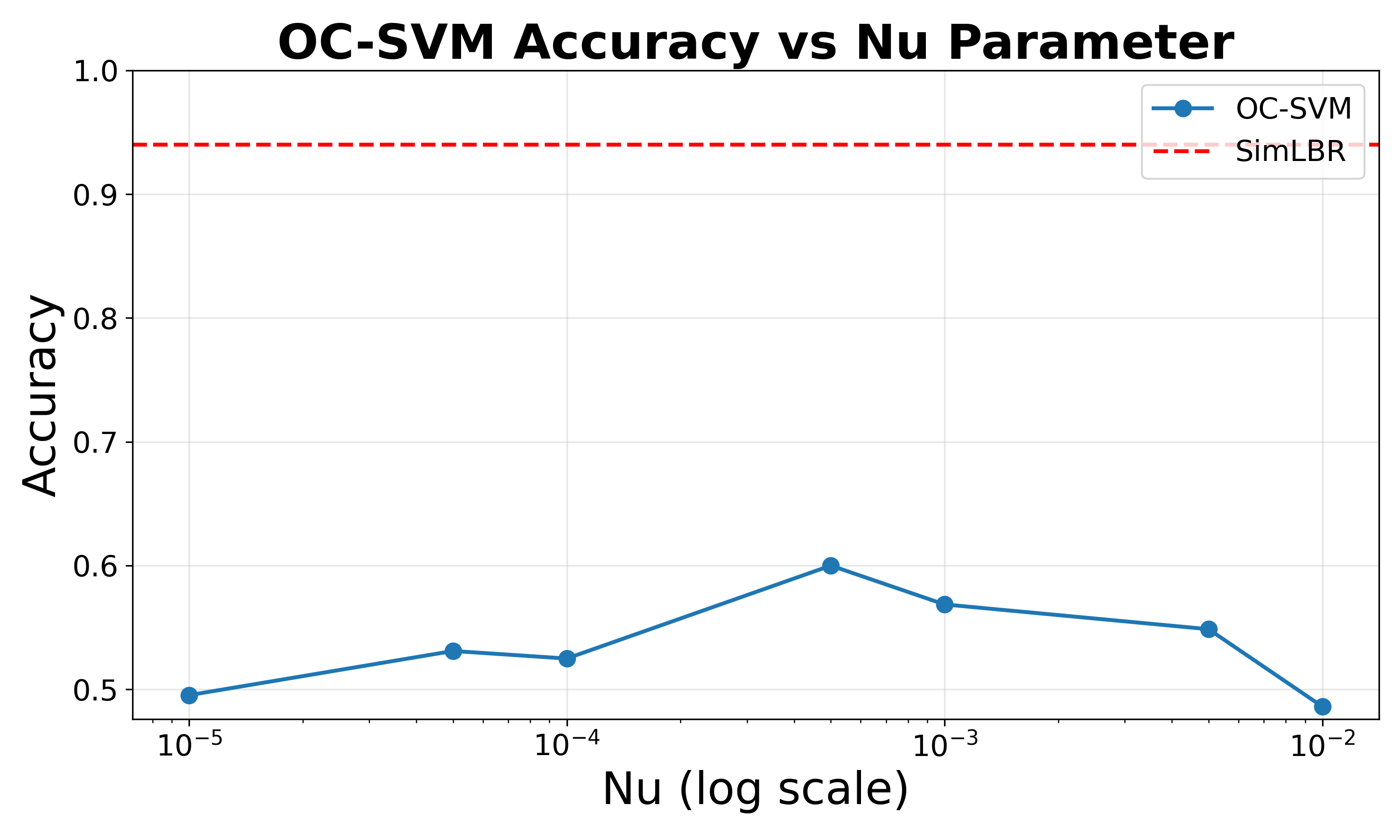}
     \subcaption{GenImage}
    \label{fig:ocsvm_genimage}
  \end{subfigure}
  \caption {\textbf{Comparison with OC-SVM:} We train a One-Class SVM using ProGAN and SD v1.4 images and evaluate on AIGC and GenImage benchmarks, respectively. SimDLR clearly outperforms the outlier detector approach when using the same latent features.}
  \label{fig:ocsvm}
\end{figure}

A natural strategy to avoid overfitting to generator–specific artifacts is to model only the real image distribution and treat AI-generated image detection as an anomaly-detection problem. To evaluate this idea, we train a One-Class SVM (OC-SVM) using DINOv3 features of real training images for both the AIGC and GenImage benchmarks. We use the \textit{RBF} kernel and do a sweep to find the best $\nu$. As shown in Figure~\ref{fig:ocsvm}, the OC-SVM performs substantially worse than SimLBR across both settings.

This result highlights a fundamental limitation of pure anomaly-detection approaches in high-dimensional image spaces. Modern generative models produce highly photorealistic images whose latent representations often lie close to those of real images. As a result, simply learning the support of the real distribution is insufficient; fake images do not reliably fall outside this region. In contrast, SimLBR introduces structured guidance by injecting controlled amounts of fake latent information into real samples during training. This guided perturbation shapes a more meaningful decision boundary around the true real-image manifold, enabling the detector to distinguish real from fake images even when fakes closely mimic real data.

\section{Additional Implementation Details}
To simplify implementation, instead of resampling different labels for the same real image during training, we iterate over the entire dataset and, whenever a fake image is encountered, we sample a real image to form a pair and assign it a fake label. This pairing strategy also ensures that all real and fake samples in the dataset are used during training.

